%% file: aaai2026.tex
\documentclass[letterpaper]{article} 
\usepackage{aaai2026}  
\usepackage{times}  
\usepackage{helvet}  
\usepackage{courier}  
\usepackage{multirow}
\usepackage{booktabs}
\usepackage{float}
\usepackage[hyphens]{url}  
\usepackage{graphicx} 
\usepackage{subcaption} 
\usepackage{natbib}  
\usepackage{caption} 
\usepackage{algorithm}
\usepackage{algorithmic}

\usepackage{newfloat}
\usepackage{listings}
\usepackage{amssymb}
\usepackage{amsmath}
\usepackage{bm}
\usepackage{xspace}
\usepackage{longtable}
\usepackage{array} 
\usepackage{calc}
\usepackage{colortbl}
\usepackage{pifont}
\usepackage{color}

\DeclareCaptionStyle{ruled}{labelfont=normalfont,labelsep=colon,strut=off} 
\floatstyle{ruled}
\newfloat{listing}{tb}{lst}{}
\floatname{listing}{Listing}
\nocopyright 

\newcommand{\paratitle}[1]{\vspace{1.5ex}\noindent\textbf{#1}}
\newcommand{\ie}{\emph{i.e.,}\xspace}

\newcommand{\eg}{\emph{e.g.,}\xspace}

\newcommand{\ignore}[1]{}

\usepackage{tcolorbox}
\definecolor{darkorange}{RGB}{255, 140, 0}
\definecolor{lightgreen}{RGB}{145, 204, 117}
\definecolor{lightyellow}{RGB}{250, 200, 88}
\definecolor{lightred}{RGB}{238, 102, 102}
\definecolor{lightblue}{RGB}{115, 192, 222}
\newtcolorbox{promptbox}[2][Prompt]{
colback=black!5!white,
arc=5pt, 
boxrule=0.5pt,
fonttitle=\bfseries,
title=#1, 
before upper={\footnotesize}, fontupper=\fontfamily{ptm}\selectfont,
colframe=#2, 
}
\usepackage{bibentry}

\begin{document}
\title{Decomposing the Entropy-Performance Exchange: The Missing Keys to Unlocking Effective Reinforcement Learning}
\author {
    Jia Deng\equalcontrib \textsuperscript{\rm 1},
    Jie Chen\equalcontrib \textsuperscript{\rm 1},
    Zhipeng Chen\textsuperscript{\rm 1},
    Wayne Xin Zhao\thanks{Corresponding author}\textsuperscript{\rm 1},
    Ji-Rong Wen\textsuperscript{\rm 1}
}
\affiliations{ %
    \textsuperscript{\rm 1}Gaoling School of Artificial Intelligence, Renmin University of China\\
    dengjia0510@outlook.com, ptyzchenjie@ruc.edu.cn, batmanfly@gmail.com
}
\maketitle

\begin{abstract}
Recently, reinforcement learning with verifiable rewards (RLVR) has been widely used for enhancing the reasoning abilities of large language models (LLMs). A core challenge in RLVR involves managing the exchange between entropy and performance of policies.
Despite the importance of this exchange, a fine-grained understanding of \textit{when} and \textit{how} this exchange operates most effectively remains limited.
To bridge this gap, we conduct a systematic empirical analysis of the entropy-performance exchange mechanism of RLVR across different levels of granularity.
Specifically, we first divide the training process into two distinct stages based on entropy dynamics, \ie \textit{rising stage} and \textit{plateau stage}, and then systematically investigate how this mechanism varies across \textit{stage-level}, \textit{instance-level}, and \textit{token-level} granularitiess.
Our analysis reveals that, in the rising stage, entropy reduction in negative samples facilitates the learning of effective reasoning patterns, which in turn drives rapid performance gains.
Moreover, in the plateau stage, learning efficiency strongly correlates with high-entropy tokens present in low-perplexity samples and those located at the end of sequences.
Motivated by these findings, we propose two methods that dynamically adjust the reward signal using perplexity and positional information to focus RL updates on tokens that exhibit high learning potential, achieving improvements compared to the baseline methods on various LLMs.
\end{abstract}

\section{Introduction}\label{sec:introduction}

Reinforcement learning with verifiable rewards (RLVR) has emerged as a key method for enhancing LLMs' reasoning capabilities, particularly in complex tasks like mathematical problem solving~\citep{DeepSeek-R1, openai2024b, still-3}. This paradigm trains models to produce more accurate reasoning chains by exploring multiple responses to each problem, then adjusting generation probabilities based on verifier-assigned rewards~\citep{zhu2025surprising}. 


Prior research has found that effective exploration constitutes the critical success factor in RLVR~\citep{nabati2025spectral,liao2025enhancing}.
Specifically, if LLMs can generate better responses during exploration, \eg responses with more rigorous logic or typical mistakes, they can better optimize their behavior, thereby achieving improved performance on downstream tasks.
To investigate a model's exploration ability and analyze its changes during the training process, existing studies consider the \textit{entropy of policy distribution} as a suitable indicator and argue that it warrants further in-depth analysis and investigation~\citep{haarnoja2017reinforcement}.
Following this idea, recent studies have shown that reducing the entropy of the policy distribution often leads to significant performance gains~\cite{cui2025entropy}, regarded as \textit{extropy-performance exchange}. 
Several works focus on directly minimizing overall entropy or each sampled instance~\cite{agarwal2025unreasonable,chen2025seed,prabhudesai2025maximizing,liu2024entropy}, while others find that targeting only the high-entropy tokens is sufficient to improve model performance~\cite{wang2025beyond,cheng2025reasoning}.


However, current investigations of the entropy-performance trade-off operate at a coarse granularity, treating RLVR training as a monolithic process. These studies primarily examine aggregate performance changes before and after training states, failing to provide a fine-grained analysis of how entropy dynamics interact with model performance throughout the training trajectory. 
In essence, RLVR training constitutes a complex learning process shaped by multiple involving elements~\cite{cui2025entropy}. These factors dynamically influence model behavior~\cite{wang2025stabilizing}, with entropy effects varying across training stages, token positions, and sampled instances—each contributing distinctively to overall performance.


Building on the above discussion, in this paper, we conduct a systematic study of the entropy-performance interplay in RLVR, revealing three key phenomena: stage-level dynamics, instance-level efficiency, and token-level significance.
Concretely, we first divide the RLVR training process into the \textit{rising stage} and \textit{plateau stage}, and find that performance improvement mechanisms  differ between these two stages. 
During the rising stage, the model primarily establishes formal reasoning patterns through  entropy reduction in negative samples. In contrast,  our analysis of the plateau stage demonstrates that tokens with significant entropy changes predominantly originate from low-PPL responses and with later position help in the final decision-making process, which highlights that different tokens possess varying learning potential.
Motivated by these insights, we propose two reward shaping techniques that dynamically reweight token advantages based on PPL and positional information, encouraging the model to focus on the tokens with the highest learning potential.
In summary, our key contributions and findings are given as follows: 
\begin{itemize}
    \item For stage-level analysis, we divide the RLVR process into a \textit{rising stage} and a \textit{plateau stage}. In the rising stage, reducing entropy in negative samples helps the model establish effective reasoning patterns. In the plateau stage, learning focuses on high-entropy tokens, leading to slower but steady gains.
    \item For instance-level analysis, we observe that tokens with significant entropy changes predominantly originate from \textit{low-perplexity responses}.
    \item For token-level analysis, high-entropy \textit{tokens at the beginning} of a response help the model explore, while \textit{tokens at the end} carry fine-grained task-specific information and assist the model in making final decisions.
    \item Based on our empirical insights, we propose two reward shaping methods by adjusting token-level advantages based on perplexity and positional information. These methods dynamically steer model updates toward tokens with higher learning potential, unlocking the potential of RLVR and leading to non-trivial performance gains across various LLMs and reasoning benchmarks.
\end{itemize}

\section{Methodology}\label{sec:methodology}

In this section, we describe the methodology used for conducting the reinforcement learning (RL) study and introduce fine-grained metrics for subsequent empirical analysis.

\subsection{The RLVR Approach}

We adopt GRPO~\citep{shao2024deepseekmath}, which is a variant specifically designed for reasoning tasks, as our core training framework.
Given the old policy \( \pi_{\theta_{\text{old}}} \) and the current policy \( \pi_\theta \), the objective function can be computed as follows:
{
\small
\begin{align}
\mathcal{J}(\theta) =& 
\mathbb{E}_{q \sim \mathcal{D},\; o \sim \pi_{\theta_{\text{old}}}} \Bigg[
\sum_{t=1}^{|o|}
\min\Big( r_t \hat{A}_t,\; 
\text{clip}(r_t,\; 1\!-\!\epsilon,\; 1\!+\!\epsilon) \hat{A}_t \Big) 
\notag \\
&\quad - \beta \cdot \mathrm{KL}\big[\pi_\theta(\cdot \mid q, o_{<t}) \,\|\, \pi_{\text{ref}}(\cdot \mid q, o_{<t})\big]
\Bigg],
\end{align}
}
where $q$ and $o$ are prompt and response sampled from the prompt dataset \( \mathcal{D} \) and the old policy \( \pi_{\theta_{\text{old}}} \), respectively. \( r_t = \frac{\pi_\theta(o_t \mid q, o_{<t})}{\pi_{\theta_{\text{old}}}(o_t \mid q, o_{<t})} \) is the importance sampling ratio, and \( \hat{A}_t \) is the estimated advantage. \( \epsilon \in \mathbb{R} \) is the clipping threshold, and \( \beta \) controls KL regularization.
GRPO redefines $\hat{A}_t$ through group-relative normalization. 

For a given prompt $q$, we sample $G$ responses, $\{o^1, o^2, \dots, o^G\}$, using $\pi_{\text{old}}$, assigning each response a binary reward $R^i$: 1.0 if correct and $-1.0$ for otherwise. 
Since the rewards are broadcast uniformly across all tokens in a response, the token-level advantage of the \( t \)-th token in the \( i \)-th response \( o^i_t \) is then computed as:
\small{
\begin{equation}
\hat{A}_t^i = \frac{R^i - \mathrm{mean}(\{R^j\}_{j=1}^G)}{\mathrm{std}(\{R^j\}_{j=1}^G)}.
\end{equation}
}

\subsection{Token-Level Metrics for RL Algorithmic Analysis} 
To enable a deeper analysis of RL algorithms in the RLVR setting, we introduce three fine-grained metrics that quantify token-level algorithmic behavior.  

\paragraph{Entropy.}
The token-level entropy $H_t$ is widely used to quantify uncertainty in the policy $\pi_\theta$'s predictions at generation step $t$~\citep{cui2025entropy,wang2025beyond}. Formally, given query $q$ and preceding tokens $o_{<t}$, it is defined over the vocabulary $\mathcal{V}$ as:
\small{
\begin{equation}
H_t = -\sum_{v \in \mathcal{V}} \pi_\theta(v \mid q, o_{<t}) \log \pi_\theta(v \mid q, o_{<t}),
\label{eq:ent}
\end{equation}
}
where $\pi_\theta(\cdot \mid q, o_{<t}) = \mathrm{softmax}(\frac{\bm{z}_t}{T})$.
Here $\bm{z}_t \in \mathbb{R}^{|V|}$ denotes the model's logits, and $T$ is the decoding temperature. Higher $H_t$ values indicate greater uncertainty in token selection, reflecting exploration potential during generation.

\paragraph{Gradient.}
To analyze how tokens drive policy updates, we estimate each token's contribution to policy updates by computing the gradient of the GRPO objective \(J_{\text{GRPO}}(o_t^i)\) with respect to the language model head layer and taking its Frobenius norm as the update magnitude proxy~\citep{wang2025stabilizing}. Formally, the Frobenius norm of the resulting gradient for the $t$-th token is computed as:
\small{
\begin{equation}
G_t = \left\| \alpha_t \,\bigl(\bm{e}(o_t) - \bm{\pi}_{\theta}\bigr)\cdot \bm{h}^\top \right\|_F,
\label{eq:grad}
\end{equation}
}
where \(\alpha_t = \hat{r}_t \cdot \min(\hat{A}_t, \text{clip}(\hat{A}_t, 1-\epsilon, 1+\epsilon))\), \(\bm{e}(o_t)(o_t^i)\in \mathbb{R}^V\) is the one-hot vector for token $o_t$, and \(\bm{\pi}_{\theta}\in \mathbb{R}^V\) is the policy distribution. \(\bm{h} \in \mathbb{R}^d\) is the output of the last transformer layer. The full derivation is in Appendix A.

\paragraph{Performance Impact.}
To quantitatively assess the impact of tokens on reasoning accuracy, we design a token replacement intervention stragety. For any token \( o_t^i \) within a generated sequence, we substitute it with the highest-probability alternative token under the current policy:
\small{
\begin{equation}
\tilde{o}_t^i = \arg\max_{v_k \in V \setminus \{o_t^i\}} \pi_{\theta}(v_k \mid q, o_{<t}).  
\end{equation}
}
Subsequent $k$ continuations are generated independently from both the original token \( o_t^i \) and the substituted token \( \tilde{o}_t^i \). The divergence in average solution accuracy between these paired continuation paths serves as a metric for the token's influence on downstream reasoning correctness:
\small{
\begin{equation}
I_t = \frac{1}{k} \sum_{j=1}^{k} \left( \text{Acc}_j(q, o_{<t}, o_t)\right) -  \frac{1}{k} \sum_{j=1}^{k}\left(\text{Acc}_j(q, o_{<t}, \tilde{o}_t) \right).
\label{eq:token_i}
\end{equation}
}
Here, Acc(·) is a binary function that returns 1 if the completed sequence leads to a correct solution, and 0 otherwise.

\section{Experimental Setup}
\label{sec:exp_setup}

\paragraph{Dataset and Benchmarks.}

For RL training, we utilize the STILL-3 dataset~\citep{still-3}, which contains 90K high-quality mathematical problems. We evaluate model performance on five established mathematical reasoning benchmarks: AIME 2024~\citep{aime2024}, AIME 2025~\citep{aime2025}, AMC 2023~\citep{amc2023}, MATH500~\citep{math500}, MINERVA~\citep{lewkowycz2022solving}, and two out-of-domain benchmarks: GPQA~\citep{rein2024gpqa} and HumanEval~\citep{chen2021evaluating}. 
For each benchmark, we report three key metrics: \textit{Acc@N} measures average accuracy across $N$ responses per query, \textit{Maj@N} evaluates majority-vote agreement among $N$ responses, and \textit{Pass@N} assesses the probability of obtaining at least one correct solution in $N$ responses. All metrics use \(N=8\) samples per problem with \(top\_p=0.95\) and temperature \(0.6\).

\paragraph{Implementation Details.}

We use \texttt{Qwen2.5-7B}~\citep{qwen2.5} and \texttt{Qwen2.5-Math-7B}~\citep{yang2024qwen25mathtechnicalreportmathematical} to conduct our experiments.
Our RL implementation builds on the verl framework~\citep{sheng2024verl} with GRPO~\citep{shao2024deepseekmath} as the core algorithm. For baseline implementation, we incorporate three key enhancements from DAPO~\citep{yu2025dapo}: clip-higher with thresholds \(\epsilon_{\text{low}}=0.2\) and \(\epsilon_{\text{high}}=0.28\), token-level policy gradient loss, and overlong reward shaping using a cache length of $1024$ and maximum response length of $8192$.
We set $\beta=0.0$ to exclude the KL divergence loss.
We employ a learning rate of \(1\text{e-}6\) with $10$-step learning rate warmup, a training batch size of $512$, and a mini-batch size of $32$—yielding $16$ gradient steps per batch.
During rollout, we set \(top\_p=0.95\) and temperature \(1.0\).
During evaluation, we generate \(8\) trajectories per prompt using nucleus sampling through \(top\_p=0.95\) and temperature \(0.6\).
All experiments run on $8\times\text{H100}$ GPUs with gradient checkpointing and BF16 precision. For Eq.~\ref{eq:token_i}, we set $k=32$.

\begin{figure}[H]
  \centering
  \begin{subfigure}[b]{0.25\textwidth}
    \centering
    \includegraphics[width=\textwidth]{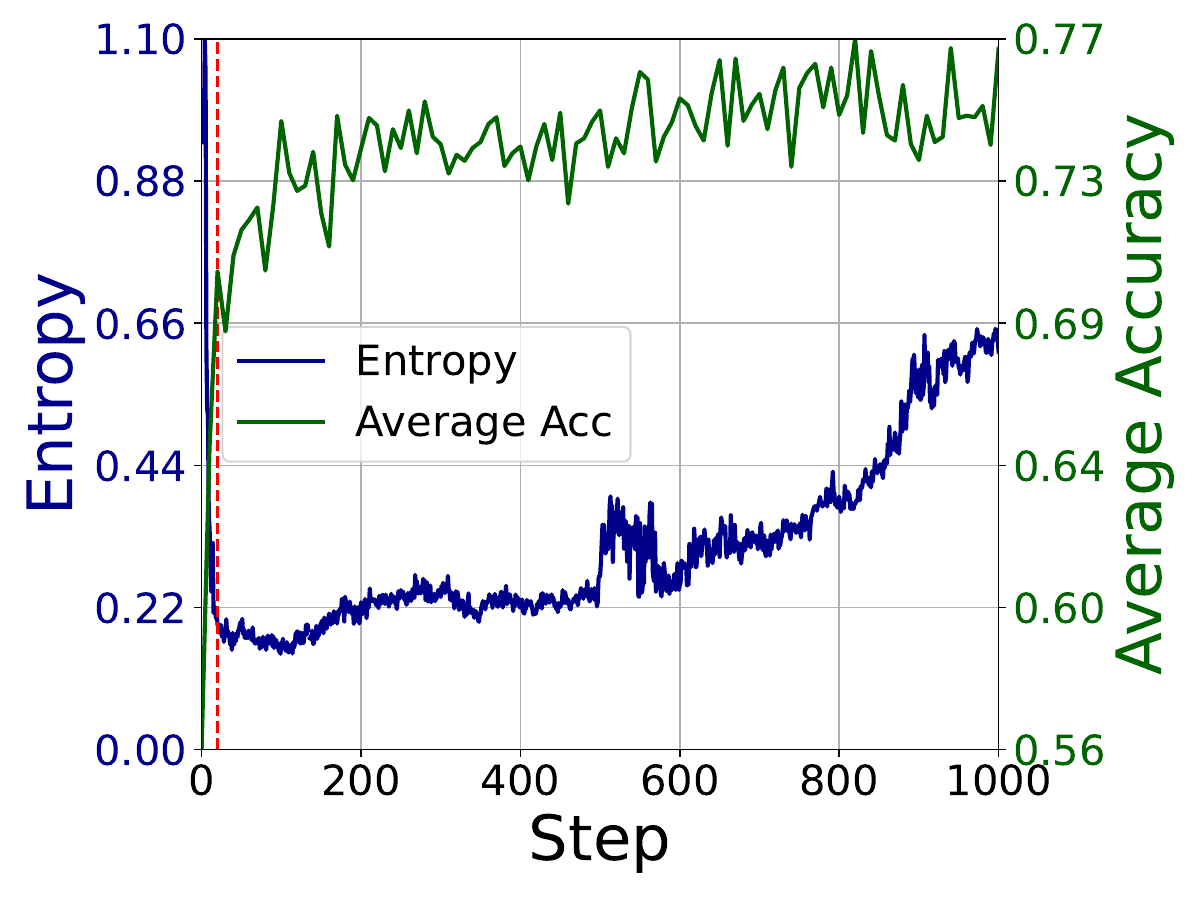}
    \caption{Entropy and accuracy trends on AIME24, AIME25 and MATH500 during GRPO training. The red line marks the transition from the rising stage to the plateau stage.}
    \label{fig:ent_and_acc}
  \end{subfigure}
  \hfill 
  \begin{subfigure}[b]{0.21\textwidth}
    \centering
    \scriptsize 
    \begin{tabular}{p{\dimexpr 0.5\linewidth-2\tabcolsep\relax} | p{\dimexpr 0.5\linewidth-2\tabcolsep\relax}}
      \hline
      \textbf{Entropy $\downarrow$} & \textbf{Frequency $\uparrow$} \\
      \hline
      Û, \_Response, ornament, pornography, enraged, tek, anz, erot, whim, Dead, ther, flirt, ĉquery ...
      & 
      \textbackslash, \textbackslash\textbackslash, (, 2, 1, =, \{, -, +, frac, \}, ), \}\{, 3, [, 0, 4, \textbackslash), ], 5, ), 6, **, 9, (\textbackslash, 8, :, sqrt, times, \_, \textbackslash\}\textbackslash, x, 7, \textbackslash[, )., cdot ...  \\
      \hline
    \end{tabular}
    \caption{The example of generated tokens with significant entropy decrease or frequency increase during the rising stage.}
    \label{tab:token-table}
  \end{subfigure}
  
  \caption{Interplay of entropy, accuracy, and token dynamics during GRPO training.
}
  \label{fig:stage_dynamics}
\end{figure}


\section{Empirical Analysis of Entropy Dynamics}

To explore the interplay between policy entropy and model performance in RLVR, we conduct a comprehensive empirical analysis examining how this relationship varies across training stages, sample quality, and token position, with all experiments based on the \texttt{Qwen2.5-7B} GRPO baseline.

\subsection{Stage-level Dynamics: Rising vs.\ Plateau Stage}

Prior work~\citep{cui2025entropy,simplerl} identifies two distinct stages in RLVR training dynamics: (1) a rapid \textit{rising stage} with quick performance improvements and decreasing policy entropy, followed by (2) a stable \textit{plateau stage} with marginal gains (Fig.~\ref{fig:ent_and_acc}). This bimodal behavior naturally raises the question: what underlying mechanisms drive performance improvements in each stage?


\paragraph{Rising Stage.}

\begin{figure}[!htbp]
  \centering
  \begin{subfigure}[t]{0.23\textwidth}
    \centering
    \includegraphics[width=\textwidth]{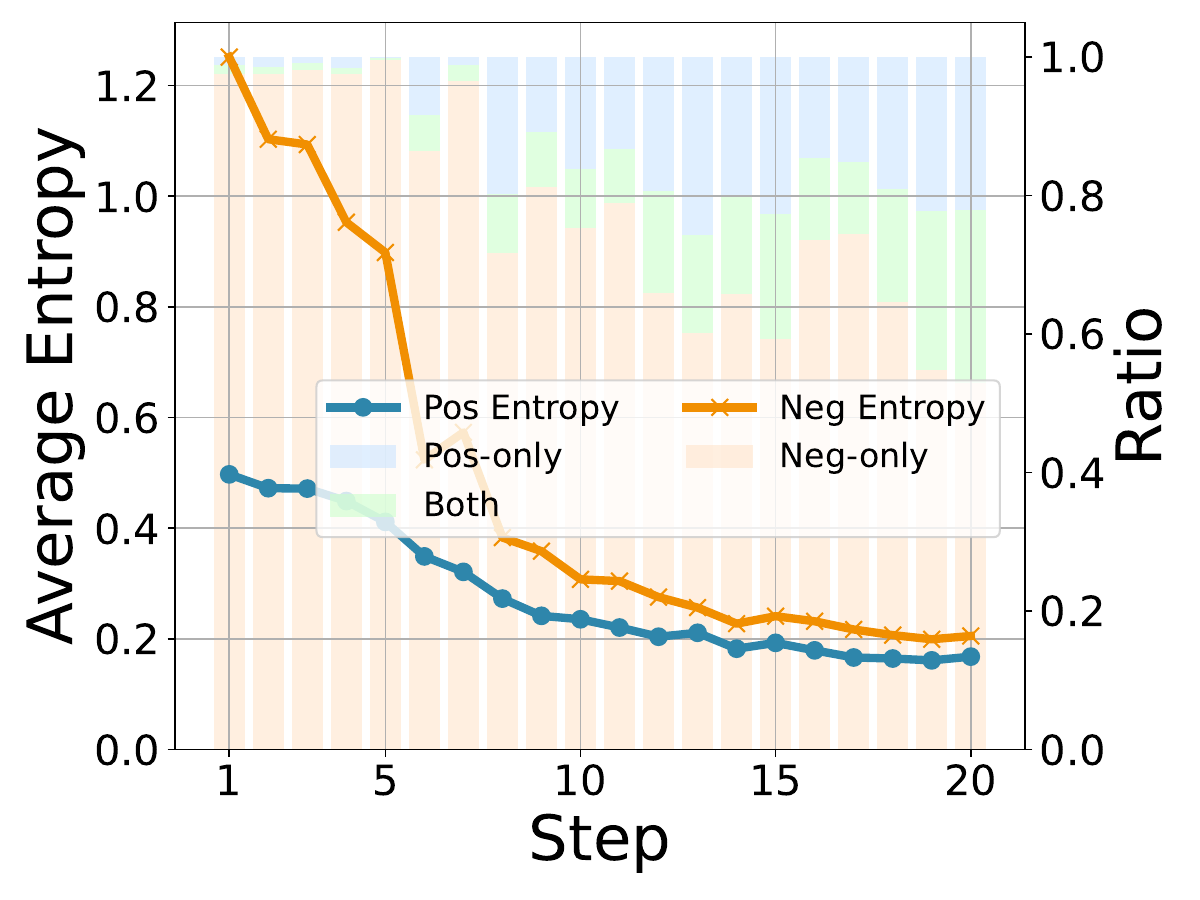}
    \caption{Entropy dynamics. The bar chart shows the sample distribution of the top 20\% tokens exhibiting the fastest entropy drop.}  
    \label{fig:pos_isu}    
  \end{subfigure}
  \hspace{0.00001\textwidth}  
  \begin{subfigure}[t]{0.23\textwidth}
    \centering
    \includegraphics[width=\textwidth]{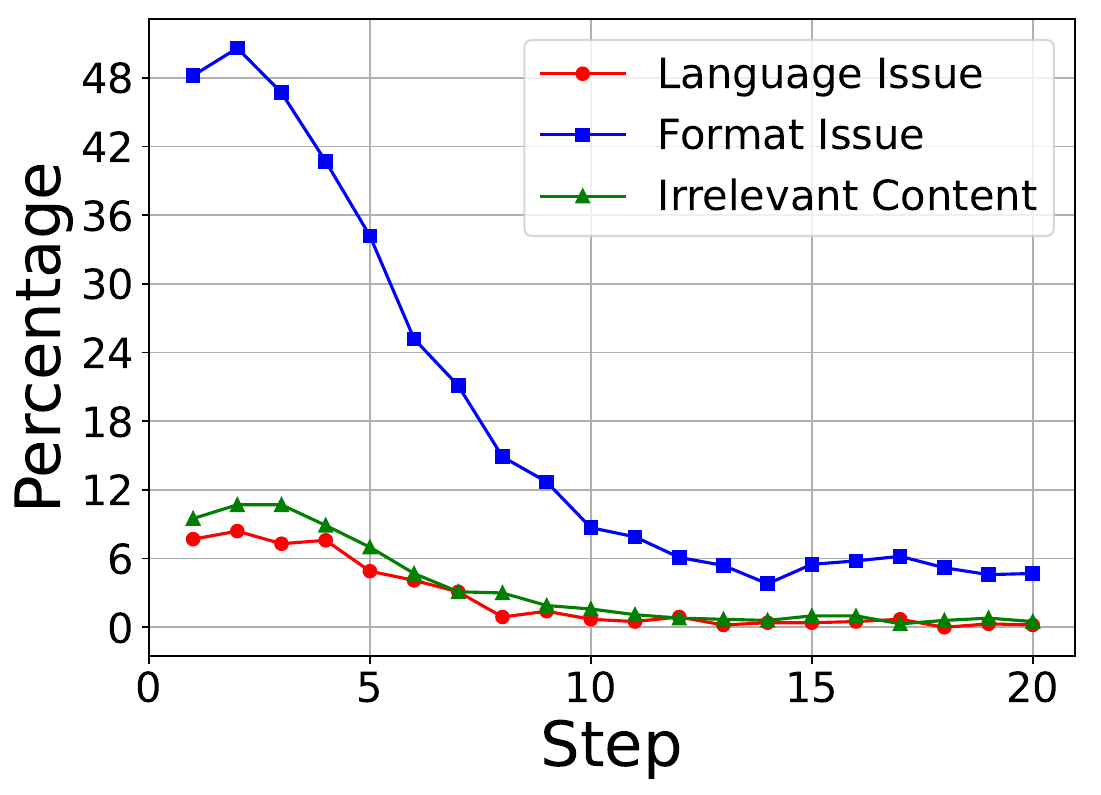}
    \caption{Proportion of model responses containing quality issues across different training steps.} 
    \label{fig:isu}  
  \end{subfigure}
  \caption{Entropy and response pattern dynamics during the rising stage.} 
\label{fig:grammar}
\end{figure}

To understand the rapid performance gains in this stage, we analyze the source of entropy reduction and its effects on model behavior. We divide the model responses at each training step into positive and negative sets, and track their entropy dynamics, revealing two main phenomena:

$\bullet$~\textbf{Entropy reduction mainly stems from negative samples.} As shown in Fig.~\ref{fig:pos_isu}, negative samples consistently exhibit higher average policy entropy than positive samples. More importantly, their entropy declines at a substantially more rapid rate during the rising stage. Also, tokens that appear exclusively in negative samples experience the fastest decline in entropy.
This suggests that penalizing incorrect reasoning paths plays an important role in the model's initial learning signal, reducing the vast space of potential errors.

$\bullet$~\textbf{Entropy reduction solidifies effective reasoning patterns.} Our analysis of token distributions (Table~\ref{tab:token-table}) reveals that the most significant entropy reductions occur in tokens unrelated to the task objective, while reasoning-critical tokens show increased frequency. As shown in Fig.~\ref{fig:isu}, this leads to a marked decrease in three key types of defective outputs: \emph{format violations} (unboxed or multiply-boxed answers), \emph{irrelevant content} (containing garbled or repetitive text), and \emph{language mixing} (multilingual responses). The details for quality assessment is detailed in Appendix B.


\paragraph{Plateau Stage.}

In this stage, as performance gains become incremental and entropy change flattens, 
we conduct a fine-grained investigation into the underlying mechanisms driving continued refinement. Specifically, we examine the distribution of token-level probability updates, analyzing both the magnitude of learning signals received by different tokens and their relationship to entropy dynamics and semantic roles.

\begin{figure}[!htbp]
  \centering
  \begin{subfigure}[b]{0.23\textwidth}
    \centering
\includegraphics[width=0.96\textwidth]{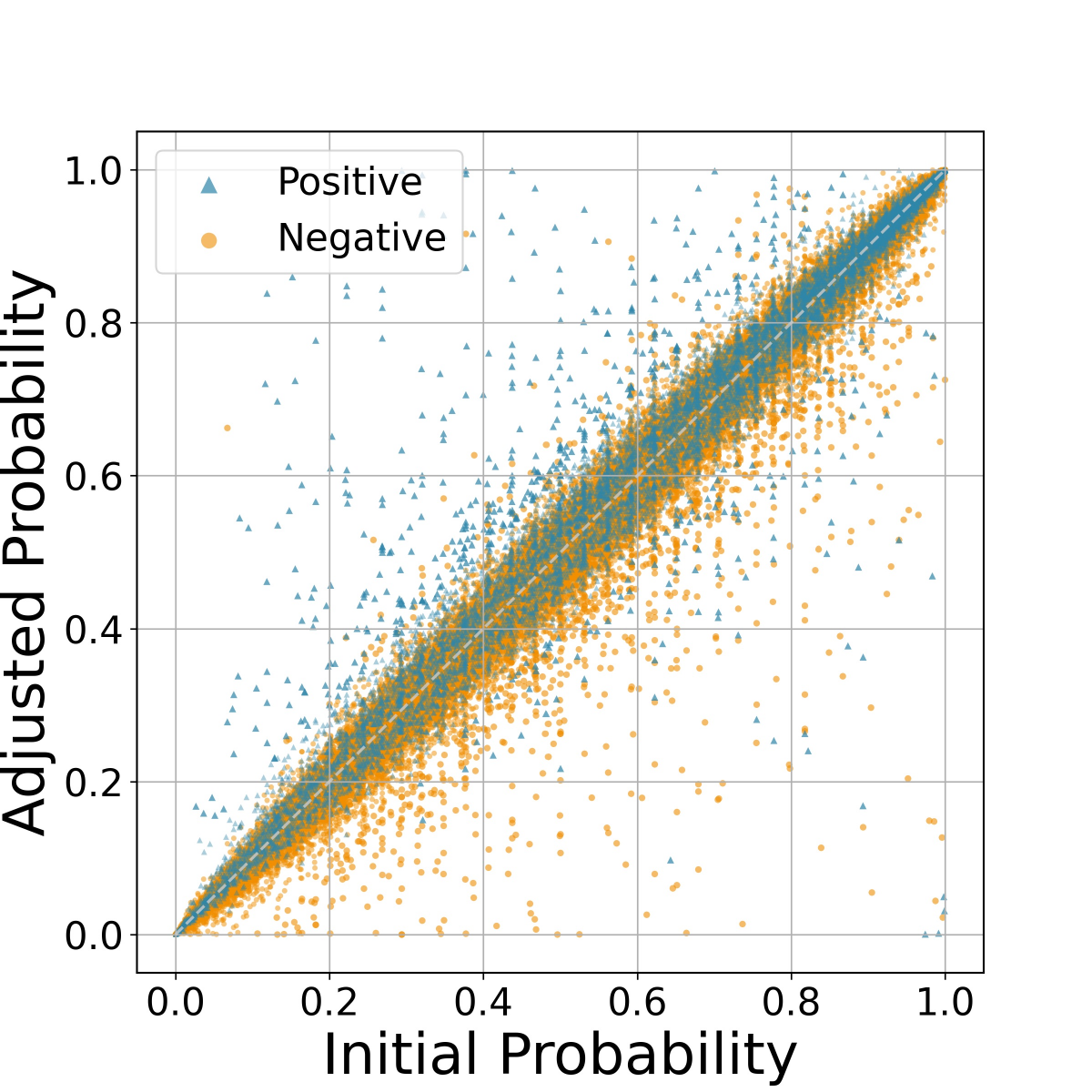}
    \caption{Token probability shifts after gradient update.}  
    \label{fig:prob_change}    
  \end{subfigure}
  \hspace{0.00001\textwidth}  
  \begin{subfigure}[b]{0.23\textwidth}
    \centering
    \includegraphics[width=\textwidth]{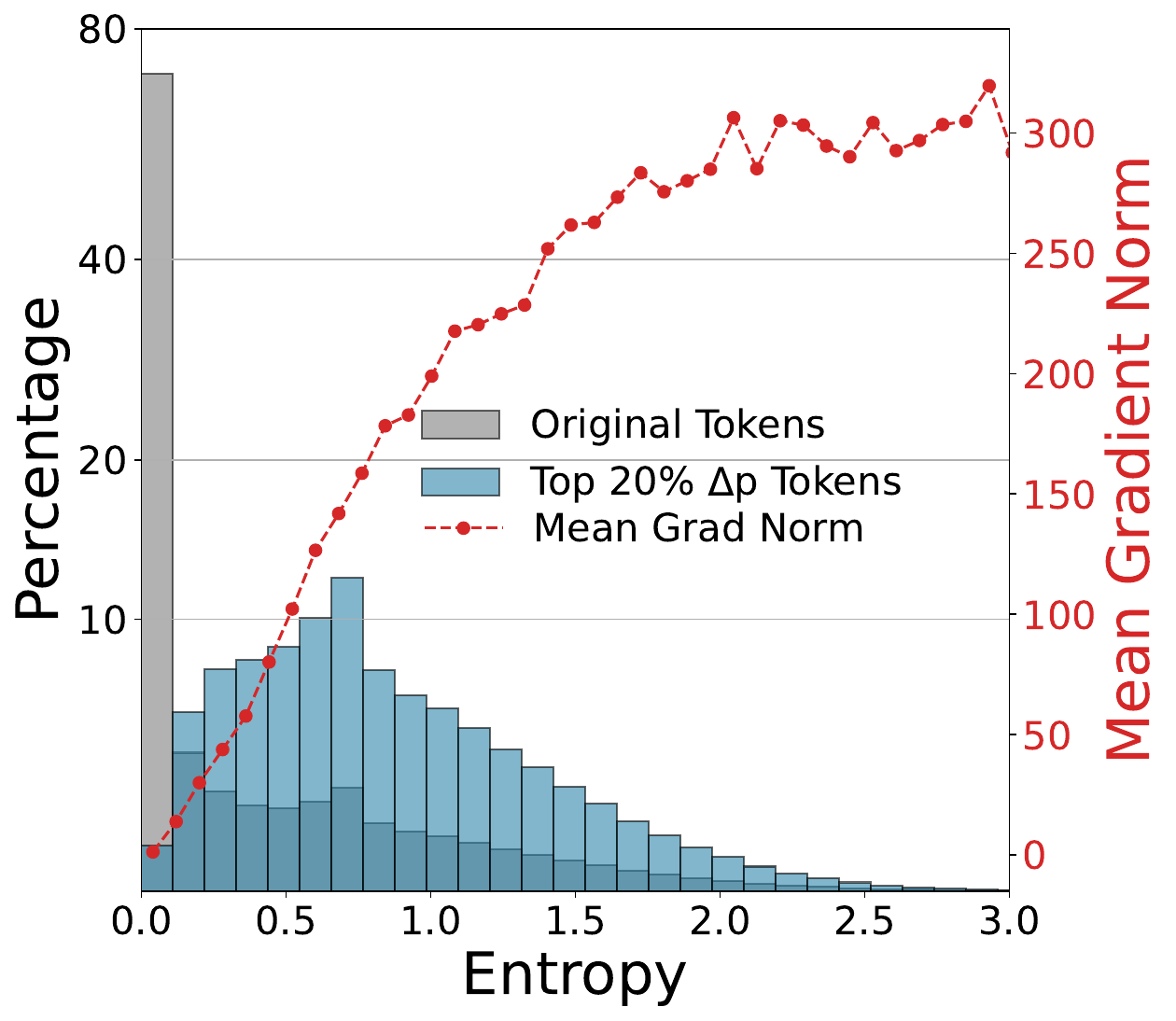}
    \caption{Token entropy and gradient distribution.} 
    \label{fig:ent_and_grad}  
  \end{subfigure}
  \caption{Token-level upate patterns.}
  \label{fig:prob_scatter}     
\end{figure}

$\bullet$~\textbf{Learning concentrates on a small subset of high-entropy, high-gradient tokens.} 
Unlike the rising stage, our analysis of token probability updates reveals that most token probabilities remain stable during the plateau stage, with over 99\% of tokens experiencing a probability change of less than 0.06 after parameter updates.
As illustrated in Fig.~\ref{fig:prob_change}, learning is instead concentrated on a small fraction of tokens where probabilities in positive samples are reinforced while those in negative samples are suppressed. In Fig.~\ref{fig:ent_and_grad}, these impactful updates primarily target high-entropy tokens. These tokens tend to produce larger gradients during backpropagation (Eq.~\ref{eq:grad}). This indicates that progress in this stage is mainly driven by resolving uncertainty at critical ``forks'' in reasoning paths~\citep{wang2025beyond}.

$\bullet$~\textbf{Updates are most sensitive for tokens associated with formal reasoning.} To further characterize these critical tokens, we categorize them by their semantic roles and analyze which types experience the largest probability changes: \emph{formal reasoning} tokens enable symbolic manipulation for computation and modeling (\eg 1, *, +); \emph{logical structuring} tokens manage the flow of reasoning (\eg but, so, however); \emph{metacognitive} tokens guide the process through self-monitoring (\eg alternatively, wait, check); and \emph{semantic support} tokens provide linguistic elements for fluency, coherence, and informativeness(\eg jump, john, she).
We provide examples of each token category in Appendix C. 
Our results show that among the top 20\% of tokens with the greatest probability updates, those associated with formal reasoning (\eg numerals, mathematical symbols) have the highest proportion ($0.039$), followed by metacognitive reasoning tokens ($0.034$), general semantic tokens ($0.033$), and logical structuring tokens ($0.031$). This targeted refinement of critical, uncertain tokens indicates a shift towards mastering the nuanced logic and precise calculations required for advanced reasoning, rather than merely reproducing structural patterns.

\subsection{Instance-level Analysis: The Role of Perplexity}

\begin{figure}[!htbp]
  \centering
  \begin{subfigure}[b]{0.23\textwidth}
    \centering
    \includegraphics[width=\textwidth]{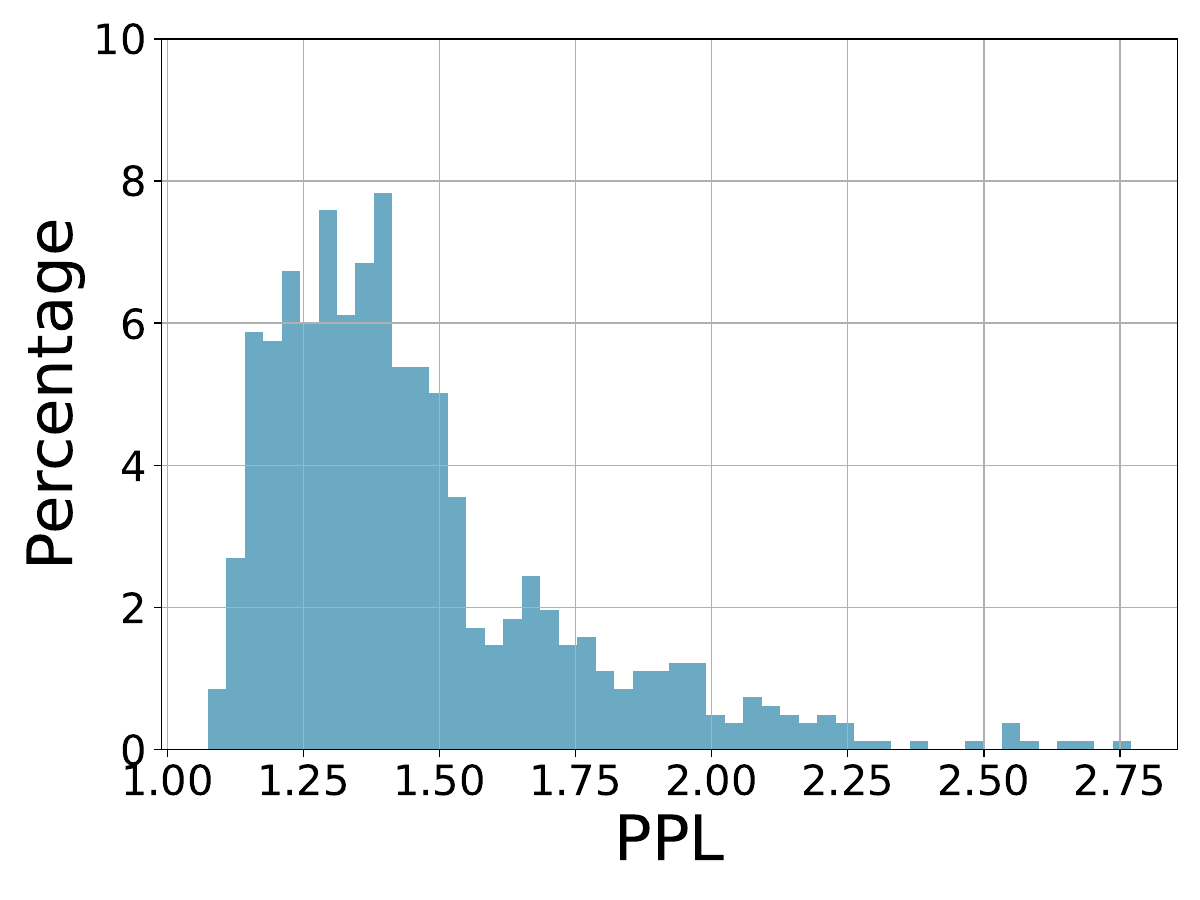}
    \caption{PPL distribution of tokens with top 20\% greatest probability shifts on the training set.}  
    \label{fig:ppl_distribution}    
  \end{subfigure}
  \hspace{0.00001\textwidth}  
  \begin{subfigure}[b]{0.23\textwidth}
    \centering
    \includegraphics[width=\textwidth]{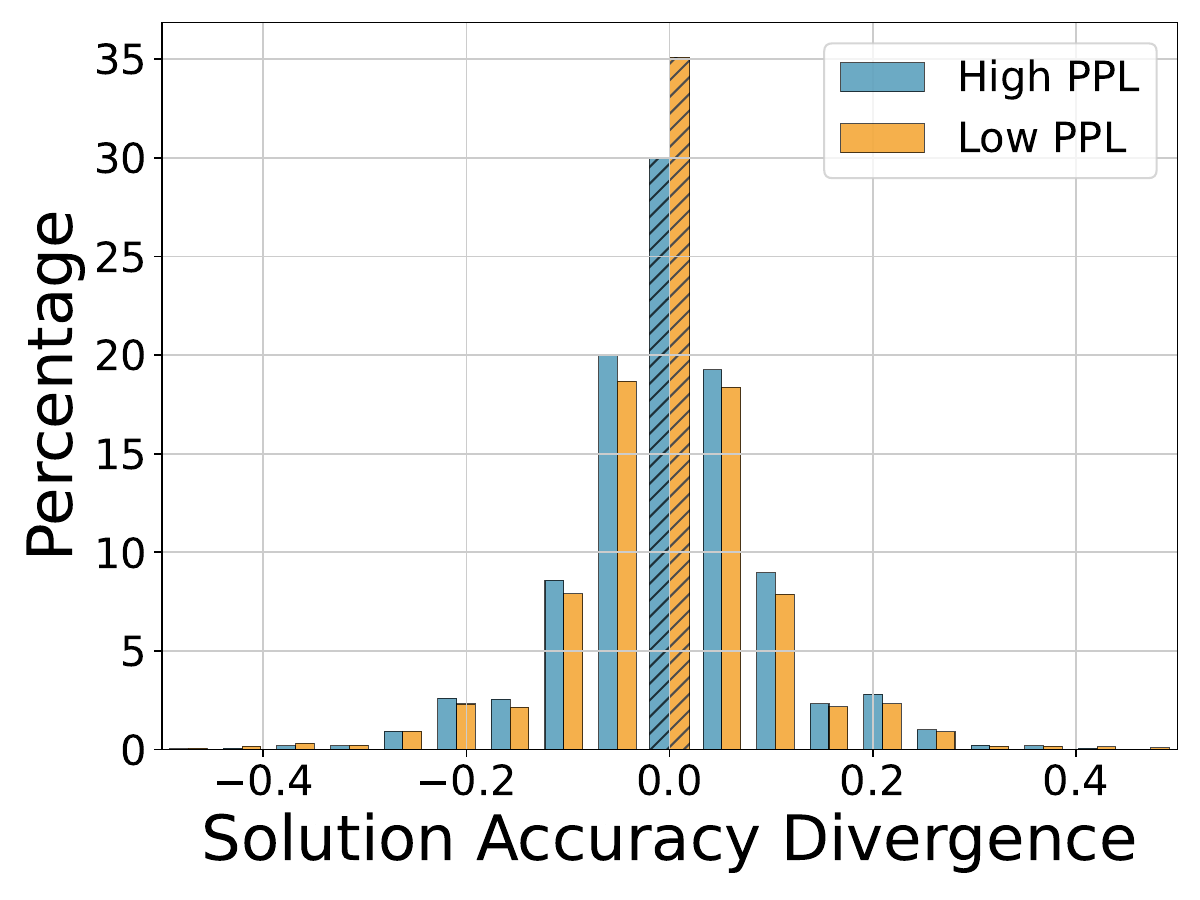}
    \caption{Accuracy changes after high-entropy token replacement in high and low PPL data.} 
    \label{fig:ppl_rollout}  
  \end{subfigure}
  \caption{Analysis of token behavior via PPL.} 
\label{fig:ppl_2}

\begin{figure}[H]
  \centering
  \begin{subfigure}[b]{0.23\textwidth}
    \centering
    \includegraphics[width=\textwidth]{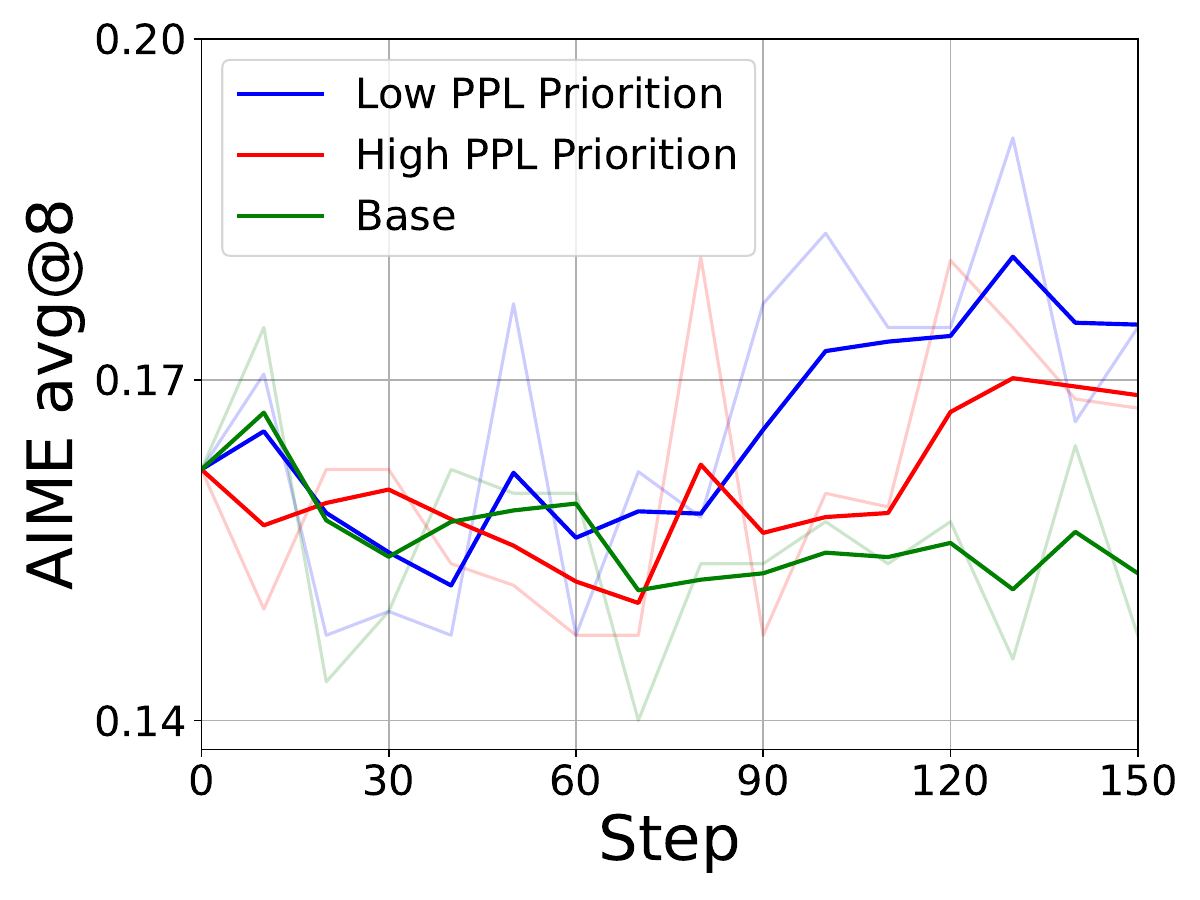}
    \caption{Accuracy.}  
    \label{fig:abli_ppl}    
  \end{subfigure}
  \hspace{0.00001\textwidth}  
  \begin{subfigure}[b]{0.23\textwidth}
    \centering
    \includegraphics[width=\textwidth]{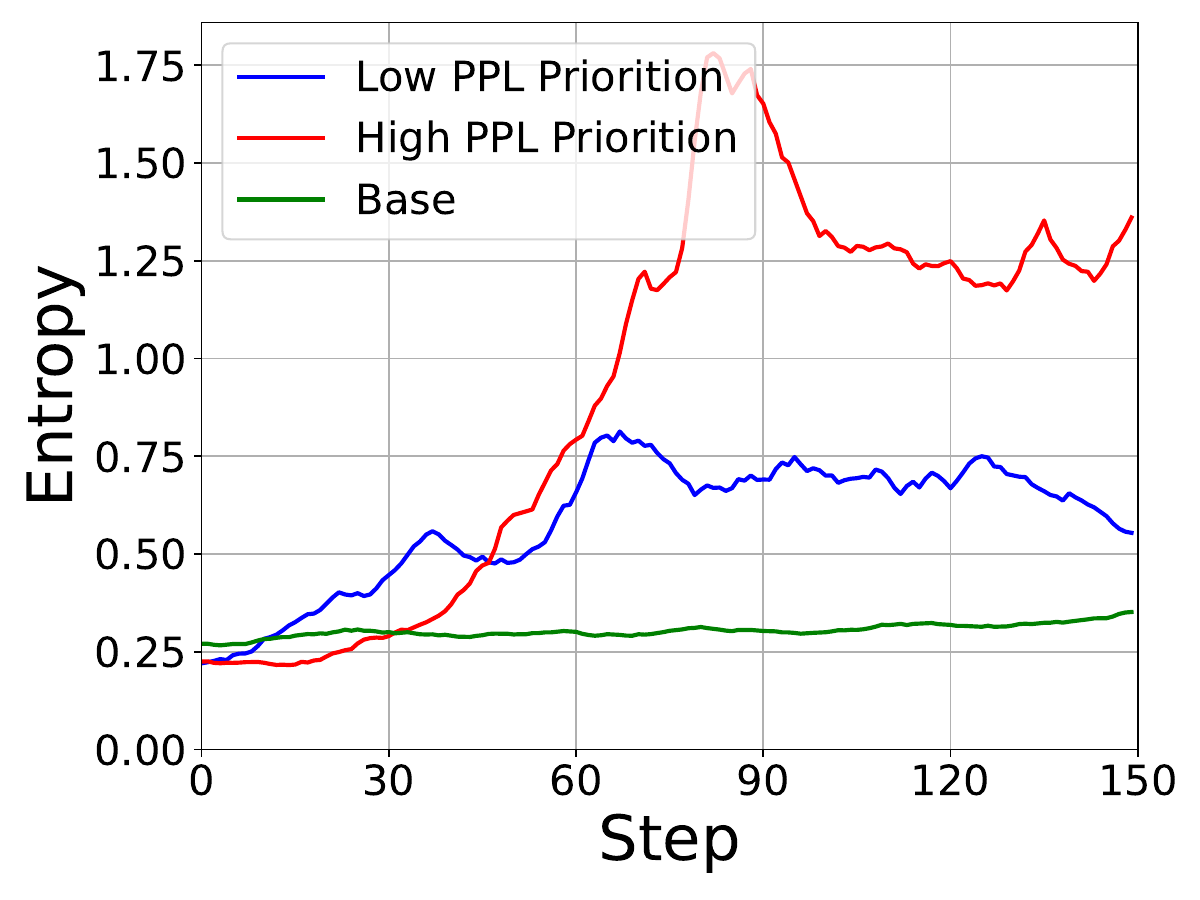}
    \caption{Entropy.} 
    \label{fig:high-low-ppl-entropy}  
  \end{subfigure}
  \caption{The effects on average accuracy on AIME24 and AIME25 and entropy by assigning higher rewards to high- or low-PPL instances with $\alpha=0.01$.} 
\label{fig:ppl_3}
\end{figure}

\end{figure}

As not all samples contribute equally to learning~\cite{chen2024allo}, to understand how instance quality affects optimization, we analyze the role of instance-level PPL, which can be regarded as a measure of the model's uncertainty over a whole sequence. 
Since low-PPL responses are generally more fluent and semantically coherent~\citep{adiwardana2020towards}, we hypothesize that these low-PPL instances are more critical for effective RLVR, which is confirmed by the following three findings from our analysis:

$\bullet$~\textbf{Learning signals are concentrated in low-PPL samples.}
To explore where learning occurs most actively, we analyze the magnitude of token probability changes during RLVR updates. As shown in Fig.~\ref{fig:ppl_distribution}, we observe a clear concentration of high-magnitude probability updates in the low-PPL region, indicating that the model's learning is more active within these generations.

$\bullet$~\textbf{Low-PPL instances represent more robust reasoning paths.} To understand the differences between samples, we apply token-level intervention analysis (Eq.~\ref{eq:token_i}) to instances sampled from both low-PPL (bottom 20\%) and high-PPL (top 20\%) groups. The results in Fig.~\ref{fig:ppl_rollout} show that replacing tokens in low-PPL responses leads to smaller changes in the final solution's accuracy compared to the same intervention in high-PPL responses, indicating that the model exhibits more robust and stable reasoning in low-PPL instances.

$\bullet$~\textbf{Prioritizing low-PPL instances enhances RLVR effectiveness.} To verify the importance of low-PPL instances, we conduct the experiment by dynamically re-weighting token advantages based on PPL. First, we compute a standardized log-PPL weight for each response $o^i$:
\small{
\begin{equation}
w_{\text{ppl}}(o^i) = \frac{\ln \mathrm{PPL}(o^i) - \mu}{\sigma}.
\label{eq:ppl_w}
\end{equation}
}
Here \( \mu \) and \( \sigma \) are the mean and standard deviation of the log-PPL values across the sampled responses for the same query \( q \), and $\alpha$ is a hyperparameter. We then compare two opposing strategies: one that adjusting the advantage with a factor of $(1 - \alpha \cdot w_{\text{ppl}}(o^i))$ of sampled instances, and another that using the factor of $(1 + \alpha \cdot w_{\text{ppl}}(o^i))$. As shown in Fig.~\ref{fig:abli_ppl}, the former one results in superior performance gains.
In contrast, focusing on high-PPL samples leads to much higher policy entropy, as shown in Figure~\ref{fig:high-low-ppl-entropy}. Further analysis of the model's generated responses on the test set reveals that this approach degrades response quality, with the frequency of responses containing quality issues rising to approximately 7\%, compared to about 3\% for the low-PPL strategy. This confirms that focusing RL updates on low-PPL samples is a more effective optimization strategy.

\subsection{Token-level Analysis: Positional Significance}

To understand how a token's effect on learning varies throughout a sequence, we analyze the interplay between token position, entropy, and optimization impact. We investigate the distribution of token entropy and importance across different positions, finding that although entropy is high at both the beginning and end of sequences, the tokens toward the end are more critical for effective RL.

\begin{figure}[!htbp]
  \centering
  \begin{subfigure}[b]{0.23\textwidth}
    \centering
    \includegraphics[width=\textwidth]{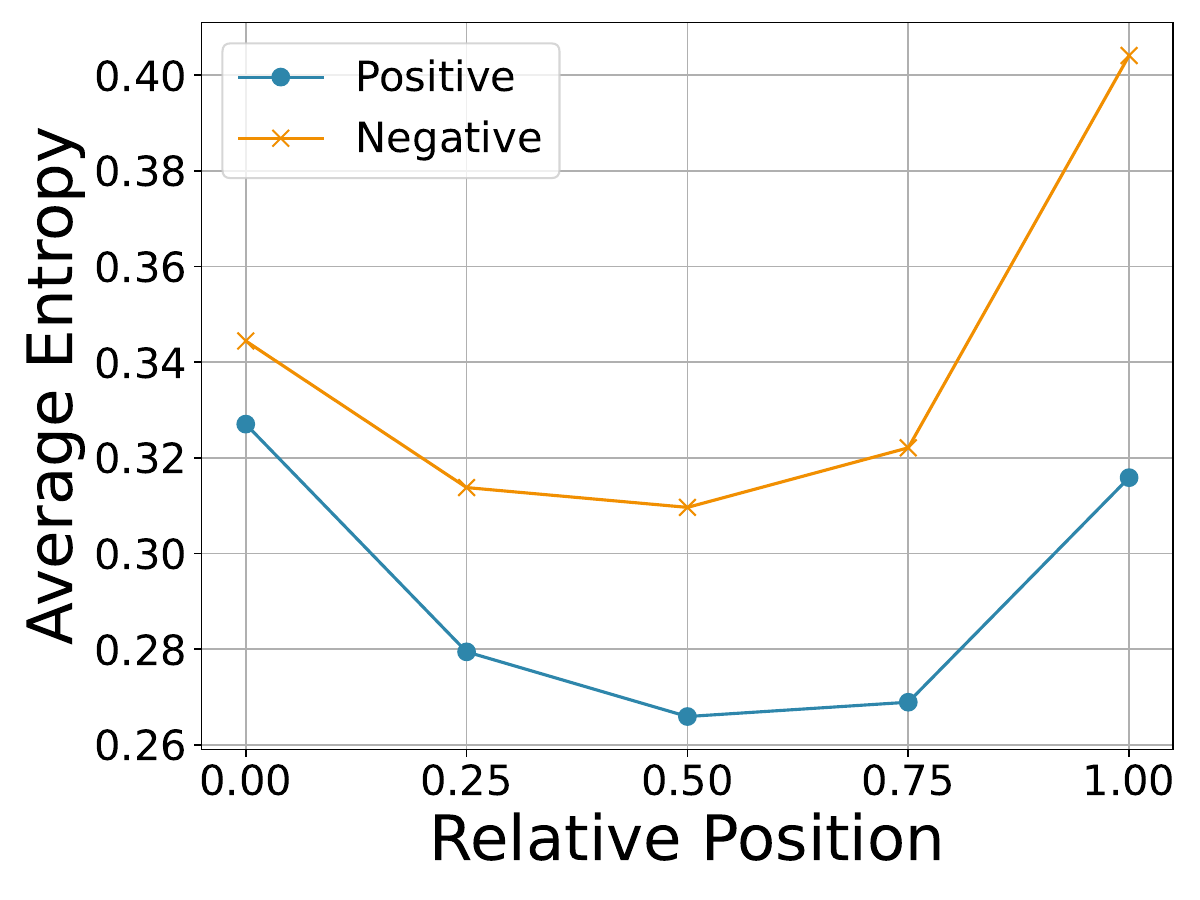}
    \caption{Positional average entropy across training data, aggregated over 1k steps.}  
    \label{fig:pos_ent}    
  \end{subfigure}
  \hspace{0.00001\textwidth}  
  \begin{subfigure}[b]{0.23\textwidth}
    \centering
    \includegraphics[width=\textwidth]{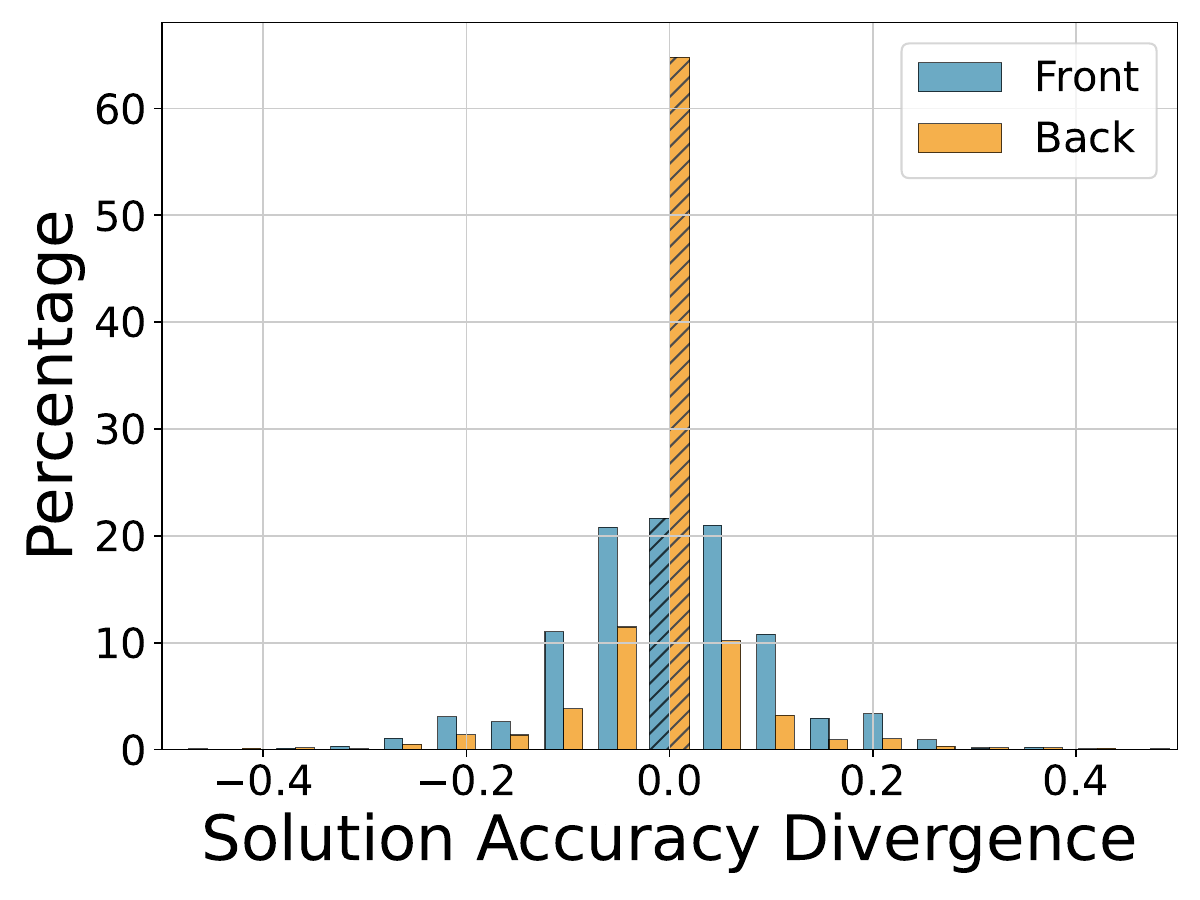}
    \caption{Accuracy shifts from high-entropy token replacement at top/bottom 20\% positions.} 
    \label{fig:loc_rollout}  
  \end{subfigure}
  \caption{Position patterns in responses.} 
\label{fig:loc_2}
\end{figure}

$\bullet$~\textbf{Token entropy follows a U-shaped distribution, with higher values at the start and end of sequences.} As illustrated in Fig~\ref{fig:pos_ent}, we observe that higher entropy concentrates at the beginning and end of a response. High entropy at the beginning reflects a broad exploration space where the model considers multiple initial approaches. In contrast, high entropy near the end of a sequence indicates uncertainty in the final decision-making process, which is directly linked to the task objective. As noted in prior work~\citep{prabhudesai2025maximizing}, there is a high correlation between model confidence in the last few tokens and overall accuracy.

$\bullet$~\textbf{Initial high-entropy tokens govern outcomes; terminal high-entropy tokens reflect reasoning  uncertainty.}
We use token-level intervention analysis in Eq.~\ref{eq:token_i} and reveal the distinct functional roles of these two high-entropy regions.
As Fig.~\ref{fig:loc_rollout} illustrates, replacing early-position tokens significantly alters the final solution's accuracy. This highlights the inherent uncertainty in the initial language space, which broadens the exploration scope and results in higher entropy. Conversely, while late-position tokens also exhibit high entropy, their minimal impact on accuracy suggests a more constrained semantic space. Interestingly, the entropy of late-position tokens in negative examples is higher than in positive ones. This subtly indicates that the model might, in the later stages of inference for incorrect solutions, implicitly detect its errors, leading to greater confusion and, consequently, elevate entropy.

\begin{figure}[t]
  \centering
  \begin{subfigure}[b]{0.23\textwidth}
    \centering
    \includegraphics[width=\textwidth]{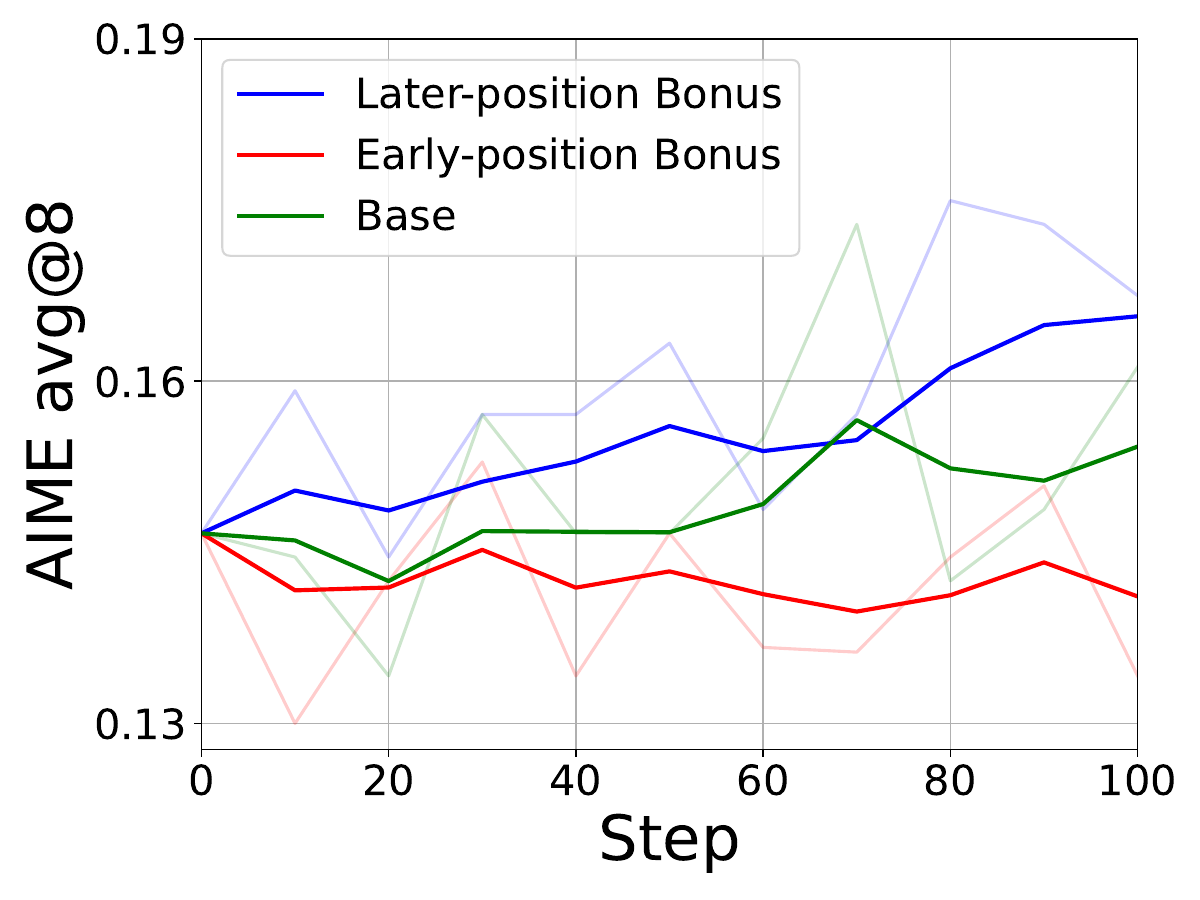}
    \caption{Accuracy.}  
    \label{fig:three_curves2}    
  \end{subfigure}
  \hspace{0.00001\textwidth}  
  \begin{subfigure}[b]{0.23\textwidth}
    \centering
    \includegraphics[width=\textwidth]{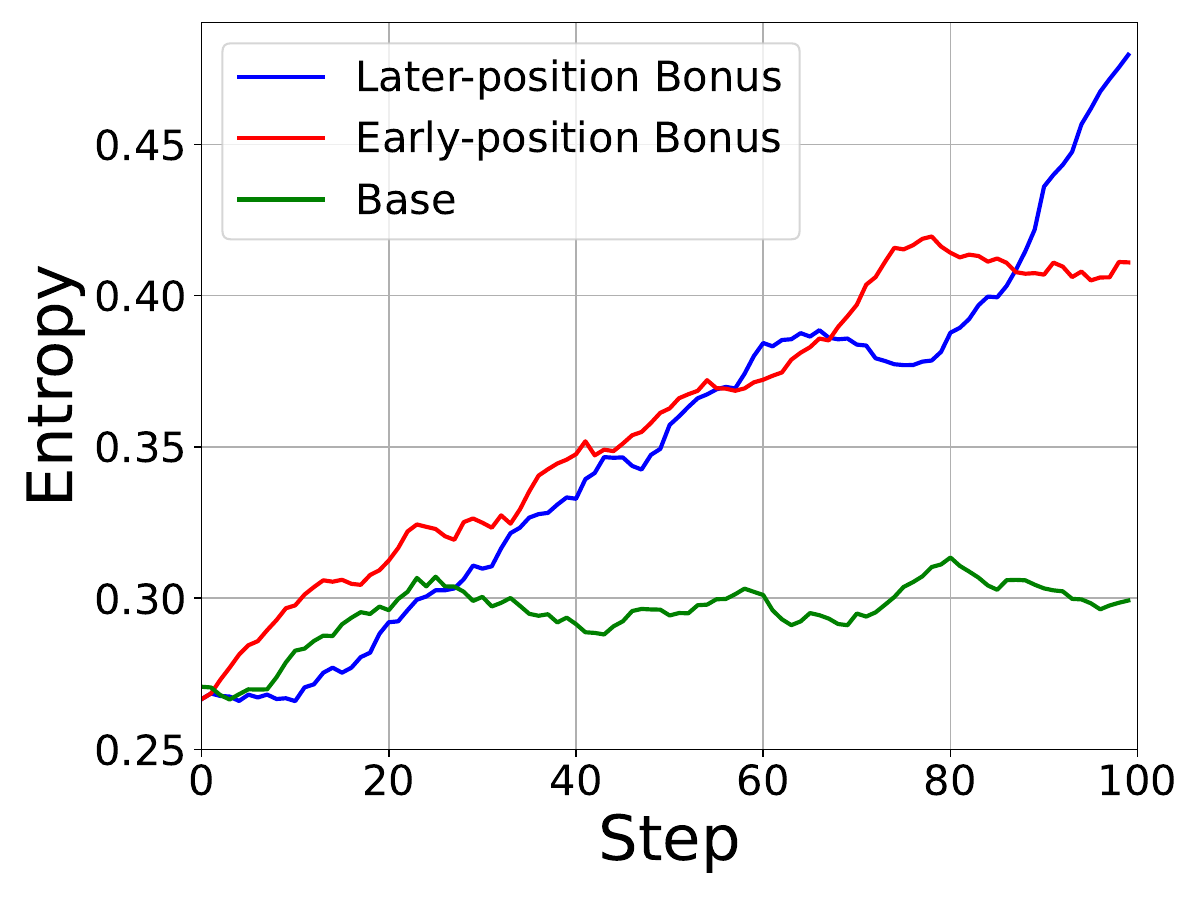}
    \caption{Entropy.} 
    \label{fig:loc_entropy}  
  \end{subfigure}
  \caption{The effects on accuracy and entropy by assigning higher rewards to different positions of responses ($\gamma=1.0$).}
\label{fig:loc_3}
\end{figure}

$\bullet$~\textbf{Optimizing tokens in later positions provides a more efficient learning signal.} To verify this, we conduct a comparative experiment by applying a positional bonus to the token advantages, defined as follows:
\small{
\begin{equation}
b^i_t = \gamma \cdot \sigma(d \cdot r^i_t).
\label{eq:loc-bonus}
\end{equation}
}
where $\gamma$ is a hyperparameter, \( \sigma \) is the sigmoid function, $r^i_t$ represents the token's relative position, and the direction parameter $d$ determines the focus of the bonus. Setting $d=1$ rewards tokens appearing later in the sequence, while setting $d=-1$ rewards tokens appearing earlier. For positive samples, this bonus is added to the original advantage to increase the reward, while for negative samples, it is subtracted to amplify the penalty. Our experiment results in Fig.~\ref{fig:three_curves2} shows that reinforcing tokens later in the sequence yields superior performance compared to both baselines with no positional bonus and the strategy that gives bonuses to early tokens.
While applying the positional bonus in either direction increases policy entropy (Figure~\ref{fig:loc_entropy}), further analysis of the generated responses reveals that rewarding early positions leads to shorter average response lengths (904 tokens) compared to rewarding later positions (1146 tokens). This suggests that optimizing the latter parts of reasoning can extend the model's reasoning time~\citep{DeepSeek-R1}, thereby improving accuracy.

\section{Advantage Shaping for Effective RLVR}
\label{sec:methods_and_results}

Drawing from our empirical analysis of entropy dynamics, we introduce two targeted methods, which are designed to steer the RLVR process by dynamically re-weighting token-level advantages, focusing learning on samples and token positions that exhibit the higher potential for efficient optimization. This section details our proposed methods, their implementation, and the experimental results demonstrating their effectiveness.

\subsection{Methods of Advantage Shaping}

We introduce two reward shaping methods designed to dynamically focus RLVR updates on parts of the generation with relatively higher learning potential. These methods re-weight token-level advantages based on sample-level perplexity and token-level position.

\paragraph{PPL-based Advantage Shaping.}

As the first strategy, we adjust token advantages to favor low-PPL samples, where learning is concentrated. For each response $o^i$ in a batch, we compute its standardized log-PPL weight $w_{\text{ppl}}(o^i)$ using Eq.~\ref{eq:ppl_w}. The advantage \( A_t \) for each token \( t \) in that response is then modulated as follows:
\small{
\begin{equation}
\tilde{A^i_t} = A^i_t \cdot \left(1 - \alpha \cdot w_{\text{ppl}}(o^i)\right).
\end{equation}
}
This method down-weights the updates from high-PPL samples, focusing the model's learning on more in-distribution reasoning paths.







\paragraph{Position-based Advantage Shaping.}

To focus optimization on the latter parts of reasoning sequences, we apply a position bonus to the token advantages. As motivated by our empirical analysis, we use the positional bonus $b^i_t$ defined in Eq.~\ref{eq:loc-bonus}. This bonus increases toward the end of the sequence and is applied based on the sign of the original advantage:
\small{
\begin{equation}
\tilde{A}^{i'}_t = A^i_t + \mathrm{sign}(A^i_t) \cdot b^i_t.
\end{equation}
}
This approach encourages the model to allocate more learning effort toward the latter parts of its reasoning process.

\begin{table*}[t]
\centering
\begin{tabular}{lcccccccccc}
\toprule
\multicolumn{1}{c}{\multirow{2}{*}{\textbf{Method}}} & \multicolumn{2}{c}{\textbf{AIME24}} & \multicolumn{2}{c}{\textbf{AIME25}} & \multicolumn{2}{c}{\textbf{AMC23}} & \multicolumn{2}{c}{\textbf{MATH500}} & \multicolumn{2}{c}{\textbf{Avg.}} \\
\cmidrule(lr){2-3} \cmidrule(lr){4-5} \cmidrule(lr){6-7} \cmidrule(lr){8-9} \cmidrule(lr){10-11}
\multicolumn{1}{c}{} & avg@8 & maj@8 & avg@8 & maj@8 & avg@8 & maj@8 & avg@8 & maj@8 & avg@8 & maj@8 \\
\midrule
\multicolumn{11}{l}{\textit{Qwen2.5-7B}} \\
\cmidrule(lr){1-11}
BASE            & 7.50  & 9.88  & 1.67  & 1.56  & 38.44 & 52.50 & 58.63 & 75.40 & 26.56 & 34.84 \\
GRPO            & \underline{21.30} & \underline{24.17} & 15.40 & 15.54 & 59.38 & 65.00 & 80.83 & 84.80 & 44.23 & 47.38 \\
GRPO+PPL        & \textbf{22.08} & \textbf{25.03} &\textbf{ 18.75} & \textbf{20.24} & \underline{60.31} & \underline{67.50} & \textbf{82.90} & \textbf{86.00} & \textbf{46.01} & \underline{49.69} \\
GRPO+POSITION   & 20.00 & 21.40 & \underline{17.08} & \underline{17.68} & \textbf{63.44} & \textbf{75.00} & \underline{81.33} & \underline{85.20} & \underline{45.46} & \textbf{49.82} \\
\midrule
\multicolumn{11}{l}{\textit{Qwen2.5-Math-7B}} \\
\cmidrule(lr){1-11}
BASE           &15.42   &20.78  & 7.50 &13.38   &42.77 &52.47 &  57.60&67.41  &30.82  &38.51  \\
GRPO            & 27.08 & 31.25 & \underline{25.00} & \underline{25.85} & 67.81 & 72.50 & \underline{86.65} & \textbf{89.00} & 51.64 & 54.65 \\
GRPO+PPL        & \underline{31.25} & \underline{37.42} & \textbf{25.42} & \textbf{26.24} & \textbf{73.44} & \textbf{82.50} & \textbf{86.73} & \underline{88.80} & \textbf{54.21} & \textbf{58.74} \\
GRPO+POSITION   & \textbf{33.75} & \textbf{39.51} & 22.92 & 24.02 & \underline{71.56} & \underline{75.00} & 86.52 & 88.20 & \underline{53.69} & \underline{56.68} \\
\bottomrule
\end{tabular}%
\caption{Results on math benchmarks across methods. Pass@k results on HumanEval and GPQA are shown in Appendix D.}
\label{tab:result-acc-highlighted}
\end{table*}


\begin{table*}[]
\centering
\begin{tabular}{lcccc} 
\toprule
\textbf{Method} & \textbf{Mean Length} & \textbf{\parbox{3.5cm}{\centering Formal Reasoning Tokens}} & \textbf{\parbox{4cm}{\centering Logical Structuring Tokens}} & \textbf{\parbox{3cm}{\centering Metacognitive Reasoning Tokens}} \\
\midrule
GRPO            & 969.06        & 501.24   & 26.31  & 0.02  \\
GRPO+PPL        & 1841.44       & 1007.07  & 44.80  & 0.18  \\
GRPO+POSITION   & 1121.28 & 607.625  & 38.10  & 0.04  \\
\bottomrule
\end{tabular}%
\caption{Comparison of average response length and token type counts in test set responses for Qwen2.5-7B.}
\label{tab:further}
\end{table*}

\subsection{Training Details}
\label{sec:train_details}

For the PPL-based reward shaping method, we apply the advantage adjustment throughout the entire RLVR training process, as PPL's measure of the model's uncertainty over a sequence is consistently applicable across the entire training period. We set the scaling hyperparameter $\alpha=0.01$.
For the positional reward shaping method, as shown in Fig.~\ref{fig:loc_entropy}, our empirical analysis reveals that applying a positional bonus can cause a rapid rise in entropy. Therefore, we apply this method selectively. The bonus is only applied during the plateau stage, beginning at step 200 and continuing for 100 steps. Also, we set a small scaling factor $\gamma=0.1$ to moderate the entropy increase. We set the bonus direction $d=1.0$. The token's relative position score $r^i_t$ is calculated as $r^i_t = m \cdot (l^i_t - n)$, where $l^i_t \in [0,1]$ is the token's relative position in the sequence, with scaling and shifting parameters $m=15$ and $n=-0.5$.




\subsection{Results and Analysis}
\label{sec:results_analysis}

We evaluate our proposed methods on  mathematical reasoning benchmarks and analyze their impact on model behavior.

\begin{figure}[t]  
\centering
\begin{subfigure}[b]{0.22\textwidth}
    \centering
    \includegraphics[width=\textwidth]{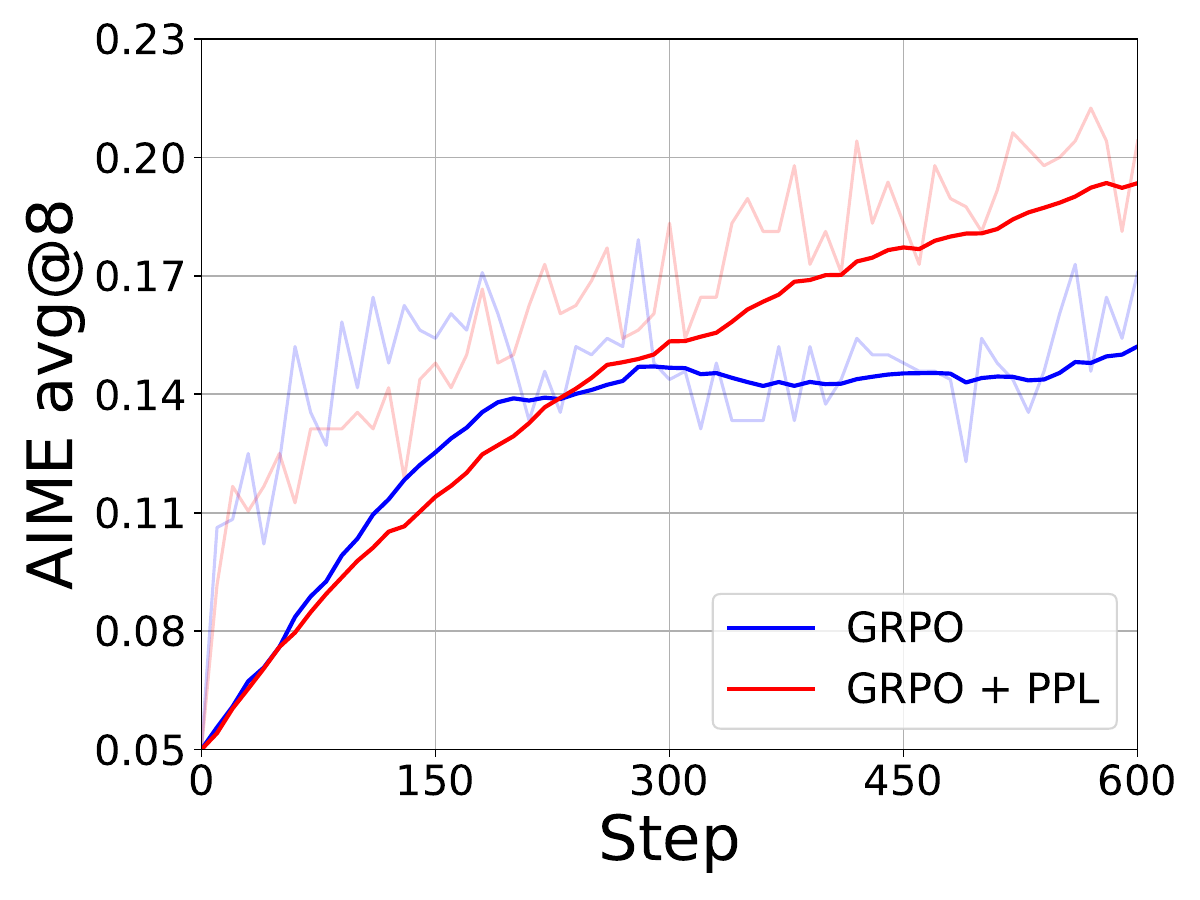}
    \caption{Qwen2.5-7B}
    \label{fig:subfig1}
\end{subfigure}
\hspace{0.00001\textwidth}
\begin{subfigure}[b]{0.22\textwidth}
    \centering
    \includegraphics[width=\textwidth]{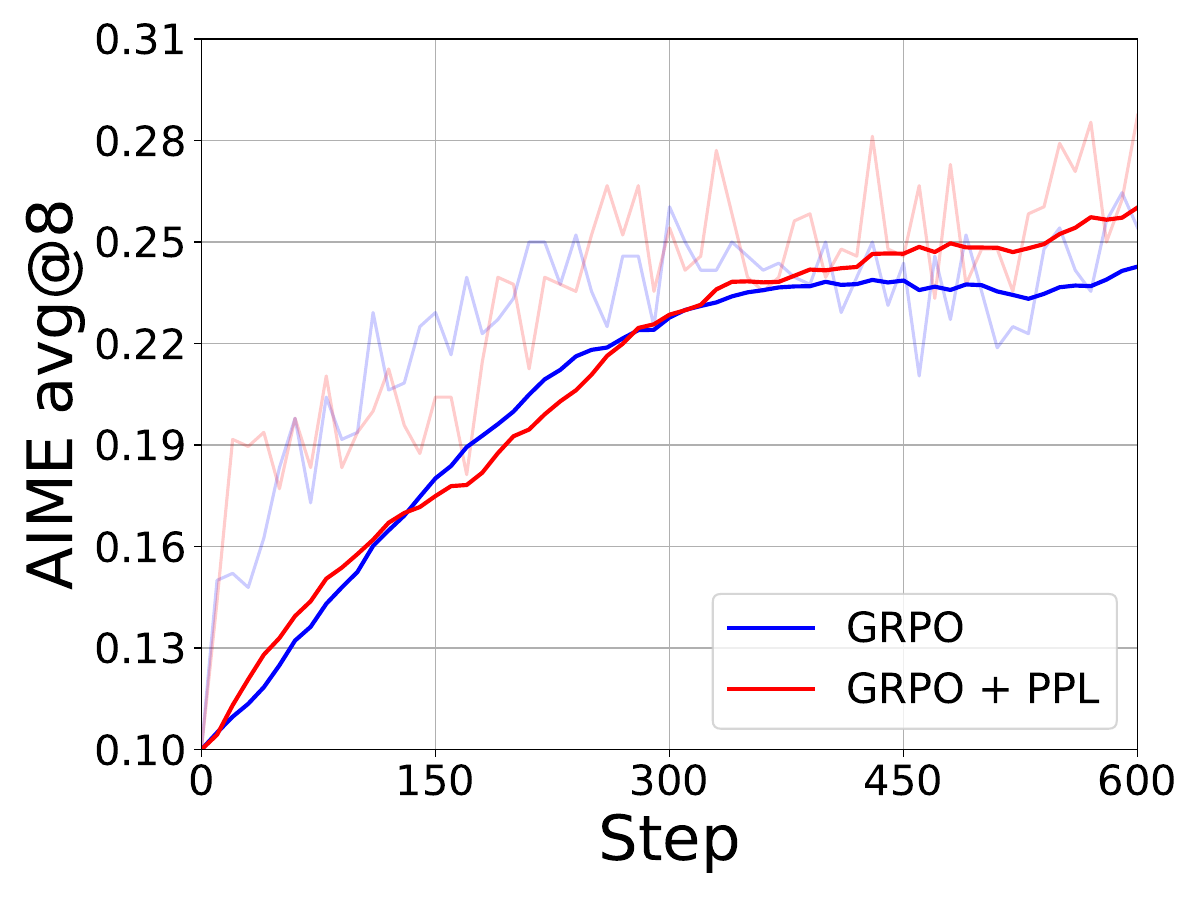}
    \caption{Qwen2.5-Math-7B}
    \label{fig:subfig2}
\end{subfigure}
\vskip 0.01cm  
\begin{subfigure}[b]{0.22\textwidth}
    \centering
    \includegraphics[width=\textwidth]{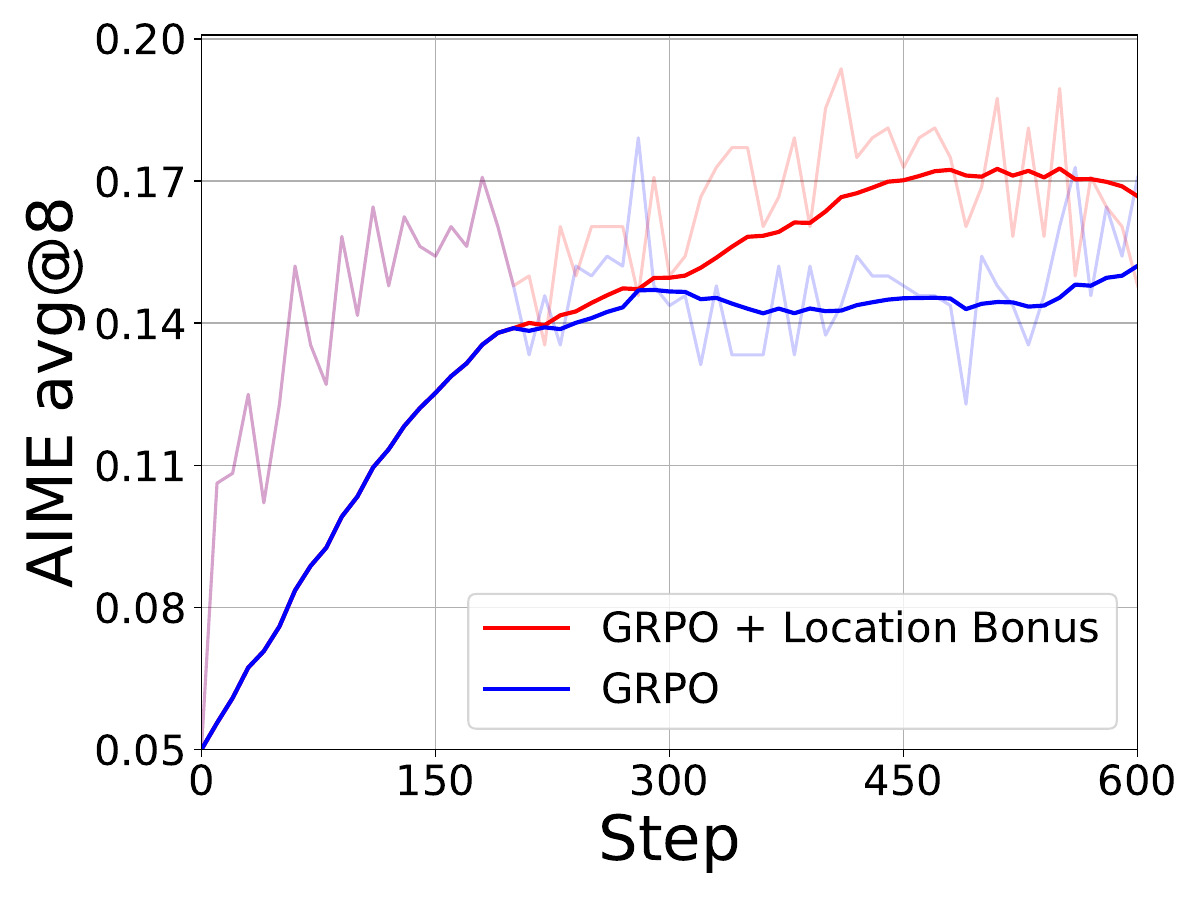}
    \caption{Qwen2.5-7B}
    \label{fig:subfig3}
\end{subfigure}
\hspace{0.00001\textwidth}
\begin{subfigure}[b]{0.22\textwidth}
    \centering
    \includegraphics[width=\textwidth]{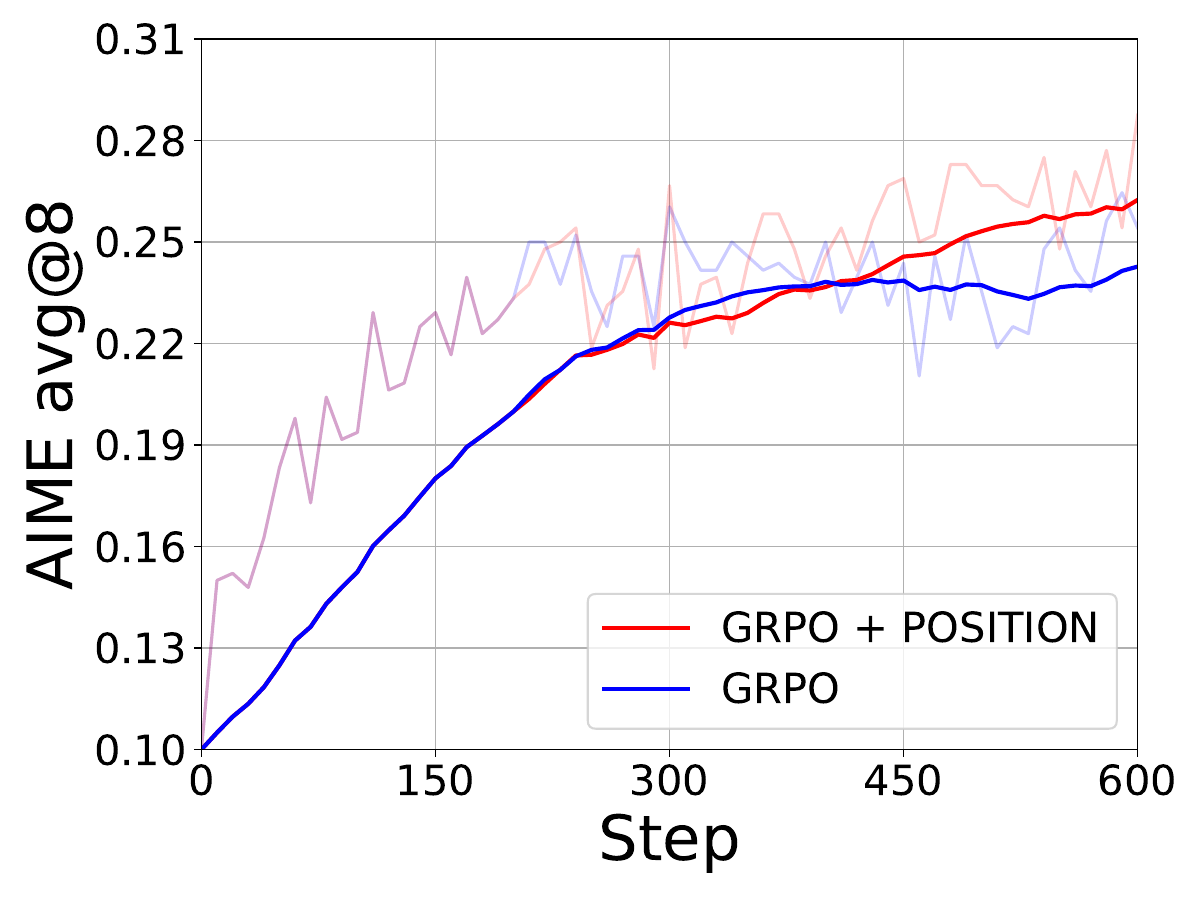}
    \caption{Qwen2.5-Math-7B}
    \label{fig:subfig4}
\end{subfigure}

\caption{Comparison of average accuracy change curves.}  
\label{fig:overall_acc}
\end{figure}

\paragraph{Overall Performance.}
As shown in Table~\ref{tab:result-acc-highlighted}, our approaches achieve subtantial improvements across the evaluation benchmarks. Compared to the GRPO baseline, they outperform it by an average of 1.51\% for the \texttt{Qwen2.5-7B} model and by 2.31\% for the \texttt{Qwen2.5-Math-7B} model, demonstrating the effectiveness of our targeted reward shaping. Moreover, our evaluations on GPQA and HumanEval reveal that both approaches exhibit enhanced generalization capabilities over the GRPO baseline.

\begin{figure}[!htbp]
  \centering
  \begin{subfigure}[b]{0.23\textwidth}
    \centering
    \includegraphics[width=\textwidth]{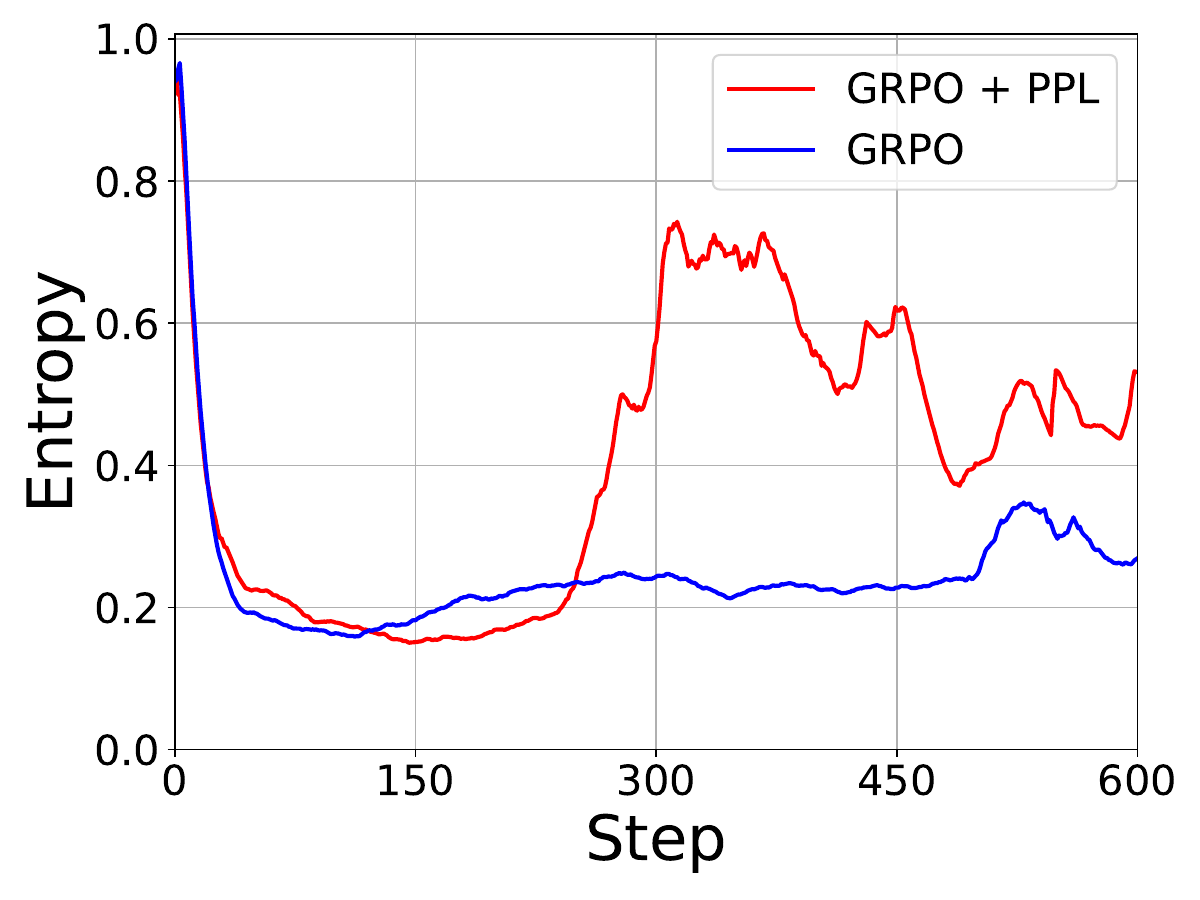}
    \caption{PPL-based  shaping.}  
    \label{fig:ent1}    
  \end{subfigure}
  \hspace{0.00001\textwidth}  
  \begin{subfigure}[b]{0.23\textwidth}
    \centering
    \includegraphics[width=\textwidth]{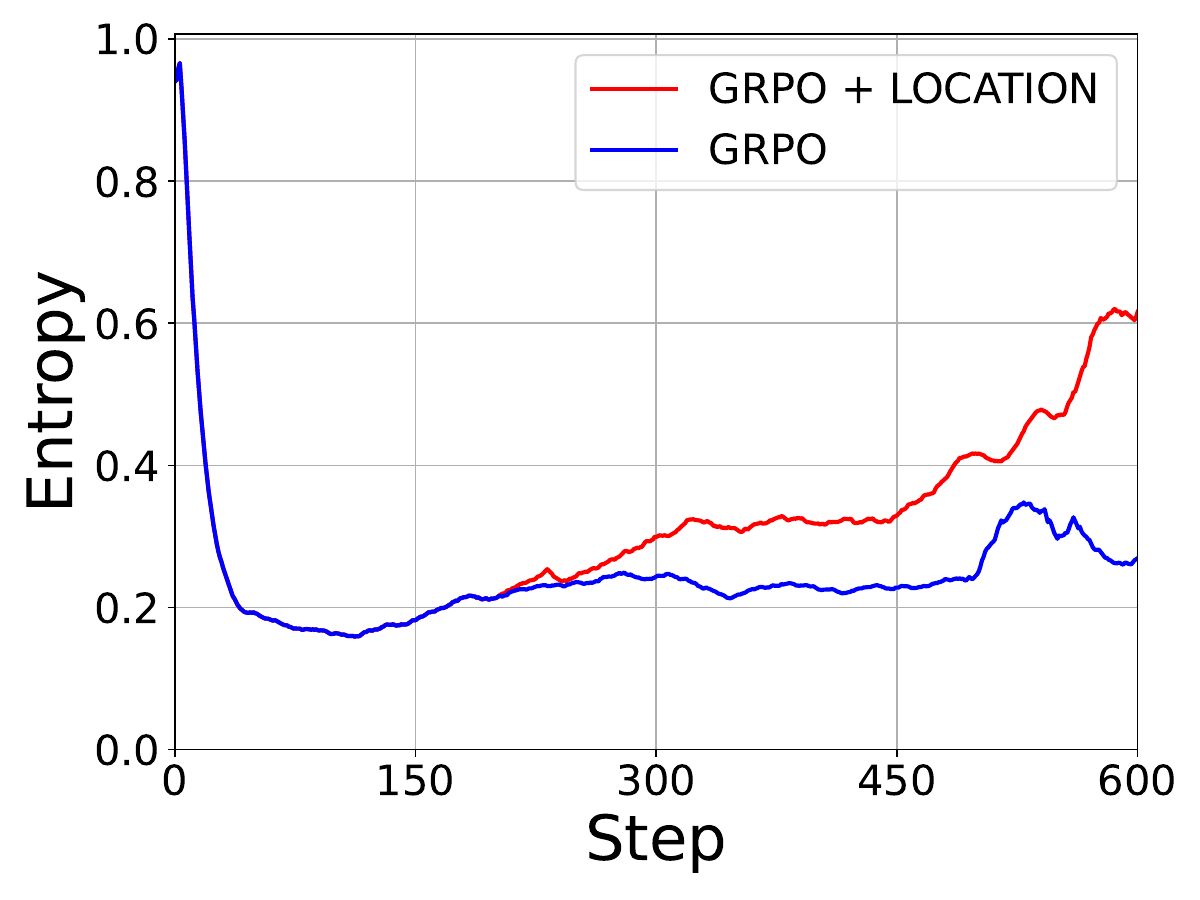}
    \caption{Position-based shaping.} 
    \label{fig:ent2}  
  \end{subfigure}
  \caption{Entropy dynamics for Qwen2.5-7B.}
\label{fig:ppl-loc-entropy}
\end{figure}

\paragraph{Entropy Dynamics.}
 As illustrated in Fig.~\ref{fig:ppl-loc-entropy}, our approaches sustain a higher level of entropy during the later stages of the plateau stage. 
 It exhibits a higher entropy trend compared to the GRPO baseline.
 This indicates that our method enables the model to retain substantial exploratory capability even in the later stages of training.

\paragraph{Response Pattern Analysis.}
We further analyze changes in response patterns by quantifying the distribution of token categories across all test sets. As shown in Table~\ref{tab:further}, both methods result in longer responses compared to the baseline, with a notable increase in tokens related to formal reasoning and logic. Formal reasoning tokens show the most significant increase, while the other categories, particularly metacognitive reasoning tokens, see smaller gains. This suggests that improving advanced cognitive abilities is inherently more difficult and may require more training steps. Case studies in Appendix E further reveal that both methods yield more detailed step-by-step breakdowns and a deeper display of the computational process compared to the baseline. Notably, the positional method encourages the model to attempt and backtrack from erroneous approaches, indicating a deeper reasoning process.



\section{Related Work}\label{sec:related_work}

\paratitle{RLVR for Mathematical Reasoning.}
Recent advancements in LLMs have highlighted the crucial role of reinforcement learning (RL) in enhancing their mathematical reasoning capabilities. Training with RL consistently improves both the accuracy and length of their responses in mathematical problem-solving~\citep{DeepSeek-R1,openai2024b}. While RL is proven to be effective, the precise mechanisms behind its success are still under investigation. A key area of focus is Reinforcement Learning with Verifiable Rewards (RLVR), a method that improves the model's ability to efficiently find correct reasoning paths~\citep{does-rl-really,wen2025reinforcement,still-3}. However, there is ongoing debate about whether RLVR genuinely fosters new reasoning patterns or simply optimizes the model's existing capabilities, with some research suggesting that distillation methods may be more effective at introducing novel reasoning strategies~\citep{does-rl-really}. Also, frameworks such as RISE have been developed to train models to address limitations like superficial self-reflection~\citep{rise}. Researchers are also focused on making RL training more efficient~\citep{accelerating-rl,rl-one}. Remarkably, some studies have shown that even 1-shot RLVR can lead to significant improvements in mathematical reasoning and self-reflection, with entropy loss playing a crucial role in promoting the necessary exploration~\citep{rl-one}.

\paratitle{Entropy of Policy Distribution in RLVR.}
Studies show that entropy plays a crucial role in enhancing model capability during RLVR~\citep{haarnoja2018soft,schulman2017proximal}. Recent work finds that a model's initial entropy level can partially predict its final performance after RL optimization~\citep{cui2025entropy}. Another analysis reveals that high-entropy tokens often correspond to key forking points in reasoning paths~\citep{bigelow2024forking}. Based on this observation, \citet{wang2025beyond} suggest that performance gains in RL are primarily driven by the learning and refinement of a small set of high-entropy tokens.
Building on this view, some approaches incorporate entropy into reward design at the token level or step level~\citep{vanlioglu2025entropy,cheng2025reasoning}, while others apply entropy regularization to maintain exploration~\citep{he2025skywork,adamczyk2025average}.
\section{Conclusion}\label{sec:conclusion}
In this paper, we presented a systematic investigation of the entropy-performance relationship in RLVR. Our analysis reveals distinct dynamics across training stages: initial performance gains emerge from entropy reduction in negative samples, while later improvements depend on reinforcing high-entropy tokens in low-perplexity contexts, particularly at sequence endings. Notably, we observe that the most significant entropy changes occur in low-PPL samples, with positional information determining their role in either exploration or precise decision-making. Building on these findings, we introduced two novel reward shaping techniques that leverage perplexity and positional information to direct RL updates toward tokens with the highest learning potential. Our methods demonstrate performance improvements across multiple reasoning benchmarks.


While our current empirical analysis and proposed methods have been validated primarily on mathematical reasoning tasks, future work will investigate their generalization to broader reasoning domains. Additionally, we plan to explore integrating our framework with existing advanced RL methods, such as DAPO, to further enhance their effectiveness.

\bibliography{aaai2026}

\input{appendix.tex}

\end{document}

%% file: appendix.tex


\lstset{%
	basicstyle={\footnotesize\ttfamily},
	numbers=left,numberstyle=\footnotesize,xleftmargin=2em,
	aboveskip=0pt,belowskip=0pt,%
	showstringspaces=false,tabsize=2,breaklines=true}
\floatstyle{ruled}
\newfloat{listing}{tb}{lst}{}
\floatname{listing}{Listing}
%
\pdfinfo{
/TemplateVersion (2026.1)
}

\setcounter{secnumdepth}{0} 

%


\definecolor{darkorange}{RGB}{255, 140, 0}
\definecolor{lightgreen}{RGB}{145, 204, 117}
\definecolor{lightyellow}{RGB}{250, 200, 88}
\definecolor{lightred}{RGB}{238, 102, 102}
\definecolor{lightblue}{RGB}{115, 192, 222}



\appendix

\section{Appendix}

\subsection{A\quad Gradient Derivation}\label{app:gradient_derivation}

We derive the gradient of the GRPO objective $J_{\text{GRPO}}$ with respect to the logits $\mathbf{z} \in \mathbb{R}^V$. Recall the policy probability for token $o_t^i$:  
\[
\bm{\pi}_{\theta}(o_t^i) = \text{Softmax}(\mathbf{z})_i = \frac{e^{z_i}}{\sum_{j=1}^{V} e^{z_j}},
\]  
where $V$ is the vocabulary size. The gradient of $\bm{\pi}_{\theta}(o_t^i)$ w.r.t. $z_k$ is:  
\[
\frac{\partial \bm{\pi}_{\theta}(o_t^i)}{\partial z_k} = \bm{\pi}_{\theta}(o_t^i) \left( \mathbb{I}(o_t^i = v_k) - \bm{\pi}_{\theta}(v_k) \right),
\]  
with $\mathbb{I}(\cdot)$ the indicator function and $v_k$ the $k$-th vocabulary token. Applying the chain rule to $J_{\text{GRPO}}$:  
\begin{align*}
\frac{\partial J_{\text{GRPO}}}{\partial z_k}
& = \left[ \hat{r}_t \cdot \min\left( \hat{A}_t, \text{clip}(\hat{A}_t, 1-\epsilon, 1+\epsilon) \right) \right] \\
& \quad \cdot \frac{1}{\bm{\pi}_{\theta}(o_t^i)} \cdot \frac{\partial \bm{\pi}_{\theta}(o_t^i)}{\partial z_k} \\
&= \alpha_t \left( \mathbb{I}(o_t^i = v_k) - \bm{\pi}_{\theta}(v_k) \right).
\end{align*}
Vectorizing over the vocabulary $V$, the gradient is:  
\begin{equation}
\frac{\partial J_{\text{GRPO}}}{\partial \mathbf{z}} = \alpha_t \left( \bm{e}(o_t) - \bm{\pi}_{\theta} \right),
\label{eq:grad_logits}
\end{equation}  
where $\bm{e}(o_t) \in \mathbb{R}^V$ is the one-hot vector for token $o_t$, $\bm{\pi}_{\theta} \in \mathbb{R}^V$ is the policy distribution, and $\alpha_t = \hat{r}_t \cdot \min(\hat{A}_t, \text{clip}(\hat{A}_t, 1-\epsilon, 1+\epsilon))$.  

Crucially, the policy update operates on the language model head weights $\mathbf{W} \in \mathbb{R}^{V \times d}$, where $\mathbf{z} = \mathbf{W} \bm{h}$ and $\bm{h} \in \mathbb{R}^d$ is the last transformer layer's output. By the chain rule:  
\[
\frac{\partial J_{\text{GRPO}}}{\partial \mathbf{W}} = \frac{\partial J_{\text{GRPO}}}{\partial \mathbf{z}} \cdot \frac{\partial \mathbf{z}}{\partial \mathbf{W}} = \underbrace{\alpha_t \left( \bm{e}(o_t) - \bm{\pi}_{\theta} \right)}_{\in \mathbb{R}^V} \cdot \bm{h}^\top,
\]  
yielding a gradient matrix $\frac{\partial J_{\text{GRPO}}}{\partial \mathbf{W}} \in \mathbb{R}^{V \times d}$. The magnitude of this update is quantified by its Frobenius norm:
\small{
\begin{equation}
G_t = \left\| \alpha_t \left( \bm{e}(o_t) - \bm{\pi}_{\theta} \right) \bm{h}^\top \right\|_F,
\label{eq:grad_final}
\end{equation} 
}
where $\|\cdot\|_F$ denotes the Frobenius norm. This serves as the token-wise update magnitude proxy.

\subsection{B\quad Methods for Detecting Low-Quality 
Responses}\label{sec:method}
We categorize low-quality responses into three types: format violations (unboxed or multiply-boxed answers), irrelevant content (garbled or repetitive text), and language mixing (multilingual responses). For format violations, we count the occurrences of ``\textbackslash\textbackslash boxed\{'' in the response string. To identify irrelevant content, we utilize Qwen2.5-32B-Instruct to determine if the response contains such content; the specific prompt used for this detection is listed in Table \ref{tab:prompt}. For language mixing, we employ a Regular Expression to check if any token's Unicode encoding falls within the range of Chinese characters.
\subsection{C\quad Token Categories in RLVR}\label{sec:token_categories}

In RLVR, tokens generated by models exhibit different functional roles that collectively drive the reasoning process. Based on their operational characteristics, we categorize tokens into four roles:

\begin{itemize}
    \item \textbf{Formal Reasoning Tokens:} Enable symbolic manipulation (\eg numbers, operators, variables, and mathematical symbols). They are essential for tasks involving structured computation or abstract modeling.
    \item \textbf{Logical Structuring Tokens:} Govern reasoning flow (\eg causal, contrastive, progressive, and parallel connectors). They help structure multi-step argumentation or explanations.
    \item \textbf{Metacognitive Tokens:} Reflect meta-cognitive functions, especially self-monitoring behaviors (\eg verifying, summarizing, and revising). These tokens actively guide the reasoning process through reflective adjustment and solution refinement.
    \item \textbf{Semantic Support Tokens:} Provide linguistic elements that ensure fluency, coherence, and informativeness (\eg core grammatical elements, domain-specific entities, and descriptive adjectives).
\end{itemize}

\noindent We provide some examples of different token categories in Table~\ref{tab:token_examples}.

\begin{table}[ht]
\centering
\begin{tabular}{l p{0.6\linewidth}}
\toprule
\textbf{Category} & \textbf{Examples} \\
\midrule
\textbf{Formal Reasoning} & Numbers (\eg `1', `3.14'), operators (\eg `+', `*', `='), variables (\eg `x', `y'), and symbols (\eg `$\pi$', `$\sqrt{2}$', `$\sum$'). \\
\\
\textbf{Logical Structuring} & Causal (\eg `therefore', `because'), contrastive (\eg `however', `but'), progressive (\eg `first', `next', `finally'), and parallel (\eg `and', `also'). \\
\\
\textbf{Metacognitive} & Verifying (\eg `Let's check'), revising (\eg `Correction', `Wait'), summarizing (\eg `In summary'), and planning (\eg `First, I will...'). \\
\\
\textbf{Semantic Support} & Grammatical elements (\eg `the', `is', `of'), domain entities (\eg `problem', `solution'), and adjectives (\eg `correct', `final'). \\
\bottomrule
\end{tabular}
\caption{Examples of Token Categories in RLVR.}
\label{tab:token_examples}
\end{table}

\subsection{D\quad Pass@k Results}\label{app:pk}
Results of pass@k on six benchmarks are shown in Tab.~\ref{tab:result-passk}. It can be seen that the average scores of our method on both the out-of-domain and in-domain benchmarks are higher than those of the GRPO baseline. However, all three methods struggle to surpass the performance of the base model on out-of-domain benchmarks, suggesting that applying reinforcement learning in the mathematics domain alone may weaken capabilities in other fields.
\begin{table*}[t]
\centering
\begin{tabular}{lccccccc}
\toprule
\textbf{Method} & \textbf{GPQA} & \textbf{HumanEval} & \textbf{AIME24} & \textbf{AIME25} & \textbf{AMC23} & \textbf{MATH500} & \textbf{Avg.} \\
\midrule
\multicolumn{8}{l}{\textit{Qwen2.5-7B}} \\
\cmidrule(lr){1-8}
BASE            & \textbf{63.13} & 14.18          & 20.30          & 7.32           & 77.50          & 87.00          & 44.91 \\
GRPO            & 48.98          & \underline{22.10} & \textbf{35.91} & \textbf{27.50} & 70.00          & \underline{90.00} & 49.08 \\
GRPO+PPL        & 50.50          & 21.49          & \underline{35.48} & 24.35          & \textbf{85.00} & \textbf{92.40} & 51.54 \\
GRPO+LOCATION   & \underline{61.11} & \textbf{23.02} & 29.75          & \underline{25.36} & \underline{80.00} & \underline{90.00} & 51.54 \\
\midrule
\multicolumn{8}{l}{\textit{Qwen2.5-Math-7B}} \\
\cmidrule(lr){1-8}
BASE            & \textbf{66.67} & \underline{26.68} & 33.76          & 21.61          & 70.96          & 84.46          & 50.69 \\
GRPO            & \underline{55.55} & 26.37          & 40.69          & \textbf{37.44} & \underline{85.00} & \underline{93.60} & 56.44 \\
GRPO+PPL        & 54.55          & \textbf{33.38} & \textbf{56.89} & \underline{33.18} & \textbf{87.50} & \textbf{94.40} & 59.98 \\
GRPO+LOCATION   & \underline{55.55} & 26.07          & \underline{55.33} & 30.85          & \underline{85.00} & 93.20          & 57.67 \\
\bottomrule
\end{tabular}%
\caption{Results for pass@k. All values are pass@8, except for humaneval which is pass@4.}
\label{tab:result-passk}
\end{table*}
\begin{table*}[htbp]
\centering
\begin{tabular}{|p{\textwidth}|}
\hline
\textbf{Prompt for Detecting Irrelevant Content} \\
\hline
Given a question and its corresponding answer, you are tasked with identifying if the response has a content redundancy issue. \textbf{Do not judge whether the answer is correct or incorrect.} Focus only on evaluating the content based on this criterion: \\
\begin{itemize}
    \item \textbf{Content redundancy}: The answer contains excessive repetition of meaningless words, or a large amount of irrelevant content, or gibberish.
\end{itemize}
Please evaluate the answer based on the criterion above. If the answer contains content redundancy, return \texttt{yes}. If it does not, return \texttt{no}. Enclose your answer in boxed.\\
\hline
\end{tabular}
\caption{Prompt for detecting irrelevant content in responses.}
\label{tab:prompt}
\end{table*}
\subsection{E\quad Case Study}\label{sec:case_stu}
We compared the answers to the same question from models trained using three different methods: GRPO, GRPO+PPL, and GRPO+POSITION. The results are presented in Tab.~\ref{tab:grpo},Tab.~\ref{tab:ppl}, and Tab.~\ref{tab:positon} respectively. We found that the responses from the GRPO+PPL and GRPO+POSITION models were noticeably more granular, with more detailed formula derivations, making them significantly easier to understand than those from the GRPO model.
\begin{table*}[htbp]  
\centering  
\begin{tabular}{|p{0.95\textwidth}|} 
\hline
Question \\ \hline
Every morning Aya goes for a $9$-kilometer-long walk and stops at a coffee shop afterwards. When she walks at a constant speed of $s$ kilometers per hour, the walk takes her 4 hours, including $t$ minutes spent in the coffee shop. When she walks $s+2$ kilometers per hour, the walk takes her 2 hours and 24 minutes, including $t$ minutes spent in the coffee shop. Suppose Aya walks at $s+\frac{1}{2}$ kilometers per hour. Find the number of minutes the walk takes her, including the $t$ minutes spent in the coffee shop. \\ \hline
Answer from GRPO \\ \hline
To determine the number of minutes the walk takes Aya when she walks at a speed of \( s + \frac{1}{2} \) kilometers per hour, we need to first set up the problem using the information given in the problem. Let's start by defining the variables and the relationships involved.The total distance of the walk is 9 kilometers, and Aya spends \( t \) minutes in the coffee shop. The total time for the walk at a speed of \( s \) kilometers per hour takes 4 hours. This total time includes both the time taken to walk and the time spent in the coffee shop. Therefore, the time taken to walk 9 kilometers at a speed of \( s \) kilometers per hour is given by:\[\text{Time taken to walk} = \frac{9}{s} \text{ hours}\]The total time taken for the walk including the time spent in the coffee shop is 4 hours. So, we have:\[\text{Time taken to walk} + \text{Time spent in coffee shop} = 4 \text{ hours}\]Therefore,\[\frac{9}{s} + \frac{t}{60} = 4 \quad \text{(1)}\]For the second scenario, where Aya walks at a speed of \( s + 2 \) kilometers per hour, the total time taken for the walk including the time spent in the coffee shop is 2 hours and 24 minutes, which is equivalent to 2.4 hours. So we have:\[\frac{9}{s + 2} + \frac{t}{60} = 2.4 \quad \text{(2)}\]Now we have two equations (1) and (2): ...\\Subtracting equation (2) from equation (1), we get:...\\
Simplifying the left side of the equation, we have:...\\Combining the fractions on the left side:...\\Multiplying both sides by \( s(s + 2) \):...\\Dividing both sides by 1.6:...\\So, we have the quadratic equation:...\\We solve this quadratic equation using the quadratic formula \( s = \frac{-b \pm \sqrt{b^2 - 4ac}}{2a} \), where \( a = 1 \), \( b = 2 \), and \( c = -11.25 \):...\\This gives two solutions:...\\Since the speed \( s \) must be positive, we discard the negative solution. Therefore,\[s = 2.5 \text{ km/h}\]
...\\Therefore, the number of minutes the walk takes Aya, including the time spent in the coffee shop, when she walks at a speed of \( s + \frac{1}{2} \) km/h is \(\boxed{204}\). \\ \hline
\end{tabular}
\caption{Answer from GRPO.}
\label{tab:grpo}
\end{table*}

\begin{table*}[htbp]  
\centering  
\begin{tabular}{|p{0.95\textwidth}|} 
\hline
Question \\ \hline
Every morning Aya goes for a $9$-kilometer-long walk and stops at a coffee shop afterwards. When she walks at a constant speed of $s$ kilometers per hour, the walk takes her 4 hours, including $t$ minutes spent in the coffee shop. When she walks $s+2$ kilometers per hour, the walk takes her 2 hours and 24 minutes, including $t$ minutes spent in the coffee shop. Suppose Aya walks at $s+\frac{1}{2}$ kilometers per hour. Find the number of minutes the walk takes her, including the $t$ minutes spent in the coffee shop. \\ \hline
Answer from GRPO+PPL \\ \hline
...\\**Step 1: Analyze the first scenario where Aya walks at a speed of \(s\) kilometers per hour.**- **Walking Time:** When Aya walks at a constant speed of $s$ kilometers per hour, the time she spends walking is the distance divided by her speed. The distance she walks is 9 kilometers.  - The walking time can be calculated as: ...\\ - **Total Time:** The total time for the walk, including the time spent in the coffee shop, is 4 hours.  - The time spent in the coffee shop is $t$ minutes, which we convert to hours by dividing by 60:    $t$ minutes = $\frac{t}{60}$ hours.  - Thus, the total time is the sum of the walking time and the time spent in the coffee shop, which equals 4 hours.  Thus,...\\Substitute the expression for the walking time into this equation:...\\Multiply through by 60 to clear the fractions:...\\**Step 2: Analyze the second scenario where Aya walks at a speed of \(s + 2\) kilometers per hour.**- **Walking Time:** When Aya walks at a speed of \(s + 2\) kilometers per hour, the walking time is the distance divided by this new speed, which is $\frac{9}{s+2}$ hours.- **Total Time:** The total time for the walk, including the time spent in the coffee shop, is 2 hours and 24 minutes. Convert 2 hours and 24 minutes to hours.  - 2 hours and 24 minutes is equal to \(2 + \frac{24}{60} = 2 + 0.4 = 2.4\) hours.  - Thus, the total time is the sum of the walking time and the time spent in the coffee shop, which equals 2.4 hours.  Thus,...\\Substitute the expression for the walking time into this equation:...\\Multiply through by 60 to clear the fractions:...\\Now we have two equations:1. $\frac{540}{s} + t = 240$2. $\frac{540}{s+2} + t = 144$Let's subtract equation (2) from equation (1):...\\Simplify the left side:...\\Combine the fractions:...\\Thus,...\\Now we solve the equation:...\\Multiply both sides by $s(s+2)$ to clear the fraction:...\\Divide both sides by 96:...\\Thus, we have:...\\This leads to the quadratic equation:...\\Multiply through by 4 to clear the fraction:...\\Rearrange the terms:...\\Now we solve the quadratic equation $4s^2 + 8s - 45 = 0$ for $s$ using the quadratic formula $s = \frac{-b \pm \sqrt{b^2 - 4ac}}{2a}$ where $a = 4$, $b = 8$, and $c = -45$.The quadratic formula is:...\\Substitute the coefficients into the formula:...\\Calculate the discriminant:...\\Thus,...\\Since $\sqrt{784} = 28$, we have two possible solutions for $s$:...\\Since speed cannot be negative, we discard $s = -\frac{9}{2}$. Thus,...\\Now that we have the value of \( s \), we substitute \( s = \frac{5}{2} \) back into one of the original equations to find \( t \). We will use Eq. (1):...\\Substitute \( s = \frac{5}{2} \):...\\Calculate \(\frac{540}{\frac{5}{2}}\):...\\Thus,...\\Solve for \( t \):...\\So, \( t = 24 \) minutes.**Step 3: Determine the time it takes Aya to complete the walk if she walks at a speed of \( s + \frac{1}{2} \) kilometers per hour.**First, we need to determine the speed \( s + \frac{1}{2} \).\( s = \frac{5}{2} \)
...\\So, the total time in minutes is:\(\boxed{204}\) \\ \hline
\end{tabular}
\caption{Answer from GRPO+PPL}
\label{tab:ppl}
\end{table*}
\begin{table*}[htbp]  
\centering  
\begin{tabular}{|p{0.95\textwidth}|} 
\hline
Question \\ \hline
Every morning Aya goes for a $9$-kilometer-long walk and stops at a coffee shop afterwards. When she walks at a constant speed of $s$ kilometers per hour, the walk takes her 4 hours, including $t$ minutes spent in the coffee shop. When she walks $s+2$ kilometers per hour, the walk takes her 2 hours and 24 minutes, including $t$ minutes spent in the coffee shop. Suppose Aya walks at $s+\frac{1}{2}$ kilometers per hour. Find the number of minutes the walk takes her, including the $t$ minutes spent in the coffee shop. \\ \hline
Answer from GRPO+Location \\ \hline
To find the value of \(s\), \(t\) and then the time Aya takes when walking at \(s + \frac{1}{2}\) kilometers per hour, we start by setting up equations based on the given information about Aya's walking speed and the total time for her walk including the time at the coffee shop.First, consider the case where Aya walks at a constant speed of \(s\) kilometers per hour. The total distance she walks is 9 kilometers and the total time for her walk including the \(t\) minutes at the coffee shop is 4 hours. The walking time at speed \(s\) can be written as \(\frac{9}{s}\) hours. Since the total time for the walk and the coffee shop visit is 4 hours, the walking time is \(4 - \frac{t}{60}\) hours. So we have the equation for the walking time at speed \(s\):\[\frac{9}{s} = 4 - \frac{t}{60}\]Next, consider the case where Aya walks at a constant speed of \(s+2\) kilometers per hour. The total time for her walk including the \(t\) minutes at the coffee shop is 2 hours and 24 minutes, which is \(2 + \frac{24}{60} = 2.4\) hours. The walking time at speed \(s+2\) can be written as \(\frac{9}{s+2}\) hours. Since the total time for the walk and the coffee shop visit is 2.4 hours, the walking time is \(2.4 - \frac{t}{60}\) hours. So we have the equation for the walking time at speed \(s+2\):...\\
Now we have a system of two equations:...\\We can now set these two expressions for \(4 - \frac{t}{60}\) and \(2.4 - \frac{t}{60}\) equal to each other:...\\ Rearrange the equation to get:...\\This equation seems problematic, so let's solve the system of equations for \(s\) and \(t\) first and then find the value of the walking time when Aya walks at \(s + \frac{1}{2}\) kilometers per hour.From the first equation, we have:...\\From the second equation, we have:...\\Now we have two equations in terms of \(s\) and \(t\):...\\We can now set the right sides of these two equations equal to each other:...\\\textbf{This simplifies to \(4 = 2.4\), which is clearly false}. So we need to solve these two equations for \(s\) and \(t\) separately.From the first equation:...\\Multiply both sides by 60 to clear the fractions:...\\Rearrange to get:...\\From the second equation:...\\Multiply both sides by 60 to clear the fractions:...\\Now we have the system of equations:...\\We now have two equations in terms of \(s\) and \(t\). We can solve these two equations for \(s\) and \(t\).First, we can equate the two expressions for \(t\) from these two equations. From equation \((1)\):...\\From equation \((2)\):...\\Now we have an equation in terms of \(s\):...\\Rearrange terms to combine the fractions:...\\Simplify the left side of the equation:...\\Simplify the expression in the numerator of the right side:...\\Now we can cross multiply to obtain:...\\Simplify the equation:...\\So now we have the equation:...\\Rearrange terms to form a quadratic equation:...\\We now have a quadratic equation to solve for \(s\). We can use the quadratic formula \(s= \frac{-b\pm \sqrt{b^2-4ac}}{2a}\), where \(a=4\), \(b=8\), and \(c= -45\). First compute the discriminant:...\\So \(\sqrt{b^2- 4ac} = \sqrt{784} = 28\). Now we can apply the formula for the roots of the equation:...\\This gives two potential solutions for \(s\):...\\Since speed \(s\) must be a positive quantity, we discard the negative value. So we have \(s= \frac{5}{2} = 2.5\) kilometers per hour\\
....\\Thus the number of minutes the walk takes her including \( t\) minutes spent in the coffee shop when she walks at \(s +\frac{1}{2}\) kilometers per hour is \(\boxed{204}\). \\ \hline
\end{tabular}
\caption{Answer from GRPO+POSITION.}
\label{tab:positon}
\end{table*}